\documentclass{article}

\usepackage{microtype}
\usepackage{graphicx}
\usepackage{subcaption}
\usepackage{booktabs} 

\usepackage{hyperref}

\usepackage[preprint]{icml2026}

\usepackage{amsmath}
\usepackage{amssymb}
\usepackage{mathtools}
\usepackage{amsthm}
\usepackage{float}
\usepackage{graphicx}
\usepackage{subcaption}
\usepackage{bbold}
\usepackage{multirow}
\usepackage{comment}
\usepackage{booktabs}
\usepackage{adjustbox}
\usepackage{algorithm}
\usepackage{algorithmic}

\usepackage[capitalize,noabbrev]{cleveref}

\theoremstyle{plain}
\newtheorem{theorem}{Theorem}[section]
\newtheorem{proposition}[theorem]{Proposition}

\theoremstyle{definition}

\theoremstyle{remark}

\usepackage[textsize=tiny]{todonotes}

\icmltitlerunning{VeFA: Vector-Based Feature Space Adaptation for Mitigating Forgetting in Fine-Tuning}

\begin{document}

\twocolumn[
  \icmltitle{VeFA: Vector-Based Feature Space Adaptation for Mitigating Forgetting in Fine-Tuning}



  \icmlsetsymbol{equal}{*}

  \begin{icmlauthorlist}
    \icmlauthor{Peng Wang}{yyy}
    \icmlauthor{Minghao Gu}{yyy}
    \icmlauthor{Qiang Huang}{yyy}
  \end{icmlauthorlist}

  \icmlaffiliation{yyy}{Department of Industrial and System Engineering, University of Southern California, Los Angeles, United States}

  \icmlcorrespondingauthor{Qiang Huang}{qianghua@usc.edu}
  \icmlcorrespondingauthor{Peng Wang}{pwang341@usc.edu}

  \icmlkeywords{Machine Learning, ICML}

  \vskip 0.3in
]



\printAffiliationsAndNotice{}  

\begin{abstract}
Catastrophic forgetting is a well-documented challenge in model fine-tuning, particularly under small-data downstream regimes. Existing parameter-efficient fine-tuning methods operate primarily in weight space, which can cause over-specialization of the model to the limited observations of the downstream data. Recent work reveals that one potential mechanism underlying such forgetting is the introduction of intruder dimensions into the representation space during fine-tuning. To mitigate overwriting of pre-trained knowledge and enhance robustness, we propose Vector-based Feature Space Adaptation (VeFA), a new fine-tuning scheme that operates directly in feature space and avoids generating intruder dimensions. VeFA applies element-wise transformation to individual feature channels, ensuring that the effective fine-tuned weights will remain in the column space of the pre-trained weight matrix. We evaluate both standard fine-tuning performance (effectiveness) and the robustness of VeFA against LoRA on multiple tasks, including image classification, NLU, and NLG benchmarks. Across these tasks, VeFA achieves comparable fine-tuning performance while consistently showing less degree of forgetting.
\end{abstract}

\section{Introduction}

Pre-trained models on large-scale datasets have demonstrated strong generalization and transferable representations across a wide range of domains and tasks \cite{bommasani2021opportunities}. However, due to distributional differences between the pre-training and downstream data, it is often necessary to fine-tune the pre-trained model. For example, a vision-language model such as CLIP \cite{radford2021learning}, pre-trained on web-scale image-text pairs, can be fine-tuned on Oxford-IIIT Pets for improved species recognition performance. Similarly, a language model such as GPT-2 \cite{radford2019language}, pre-trained on broad web text, can be fine-tuned on WebNLG \cite{gardent2017webnlg} for data-to-text generation. This pre-training-to-fine-tuning paradigm enables efficient knowledge transfer and is especially beneficial when downstream labels are scarce.

Fine-tuning is typically achieved by adjusting the pre-trained model in the weight space to better align with the distribution and characteristics of the downstream data \cite{devlin2019bert}. Traditional methods include full fine-tuning and linear probing \cite{kumar2022fine}. Full fine-tuning updates all the parameters of the pre-trained model using the downstream task data. Linear probing updates only the final linear layer of the frozen pre-trained model. To balance adaptation capability and efficiency, parameter-efficient fine-tuning (PEFT) methods have received increasing attention in recent years \cite{zaken2021bitfit,houlsby2019parameter,hu2022lora}. Among PEFT methods, LoRA \cite{hu2022lora} injects low-rank update matrices into frozen weights and achieves strong downstream adaptation. These PEFT approaches exceed the performance of linear probing while avoiding the computational and overfitting costs of full fine-tuning.

However, current fine-tuning methods primarily adapt the pre-trained model in the weight space and allow the learned representations to exit the column space induced by the pre-trained weight matrix.  Intruder dimensions \cite{shuttleworth2024lora} can potentially be introduced during fine-tuning, adding new directions that are not present in the original pre-trained column space. Although this subspace escape increases expressivity and enables a tighter fit to the downstream data, it risks overwriting the general representations acquired during pre-training. As a result, this can lead to overfitting and catastrophic forgetting, particularly when downstream data are limited—a phenomenon commonly referred to as {catastrophic forgetting} \cite{french1999catastrophic, kirkpatrick2017overcoming, kemker2018measuring, serra2018overcoming}. \emph{The goal of this work is to constrain adaptation under the small-data downstream regime so that the model can exploit the additional expressivity of fine-tuning, while explicitly mitigating overfitting to the limited downstream samples and preventing forgetting of the broadly generalizable representations acquired during pre-training}.

Weight-space fine-tuning methods typically assume that the input–output mapping learned during pre-training is insufficiently aligned with the downstream tasks. Consequently, they replace the pre-trained weights $\boldsymbol{W}_0$ with adapted weights $\boldsymbol{W}_{\mathrm{ft}}$ to achieve effective adaptation. However, the discrepancy between the pre-training dataset and the downstream dataset is often unknown or unmeasurable. Fine-tuning the model in its weight space using \emph{limited downstream data} inevitably modifies partial model parameters, and therefore risks the loss of knowledge learned during the pre-training stage (forgetting). \emph{Essentially, the fundamental question is how to represent and integrate the unobservable downstream data discrepancy into the pre-trained model}. 

Instead of modifying weights in the weight space, we propose to fine-tune models in the feature space. The key insight is to constrain the adaptation at each layer so that the fine-tuned model always remains within the column space of the pre-trained $\boldsymbol{W}_0$. This guarantees that downstream updates cannot drift away from the representational subspace established during large-scale pre-training, thereby better preserving the broad knowledge already encoded. Although such feature-space adaptation nominally reduces the expressive power compared to unconstrained weight-space adaptation, the over-parameterization of modern neural networks ensures that there is still ample capacity to fit downstream tasks \cite{kirkpatrick2017overcoming}, enabling fine-tuned models to achieve comparable performance while exhibiting consistently stronger robustness. This design is further motivated by the fact that pre-training typically involves data that are orders of magnitude larger and more diverse than any downstream dataset, providing the pre-trained model with strong generalization and even zero-shot capability. Fine-tuning should therefore preserve and leverage the structure embedded in $\boldsymbol{W}_0$ rather than overwrite it, thereby ensuring stability while still accommodating downstream distributional shifts.

\begin{figure}[htbp]
    \centering
    \includegraphics[width=0.4\textwidth]{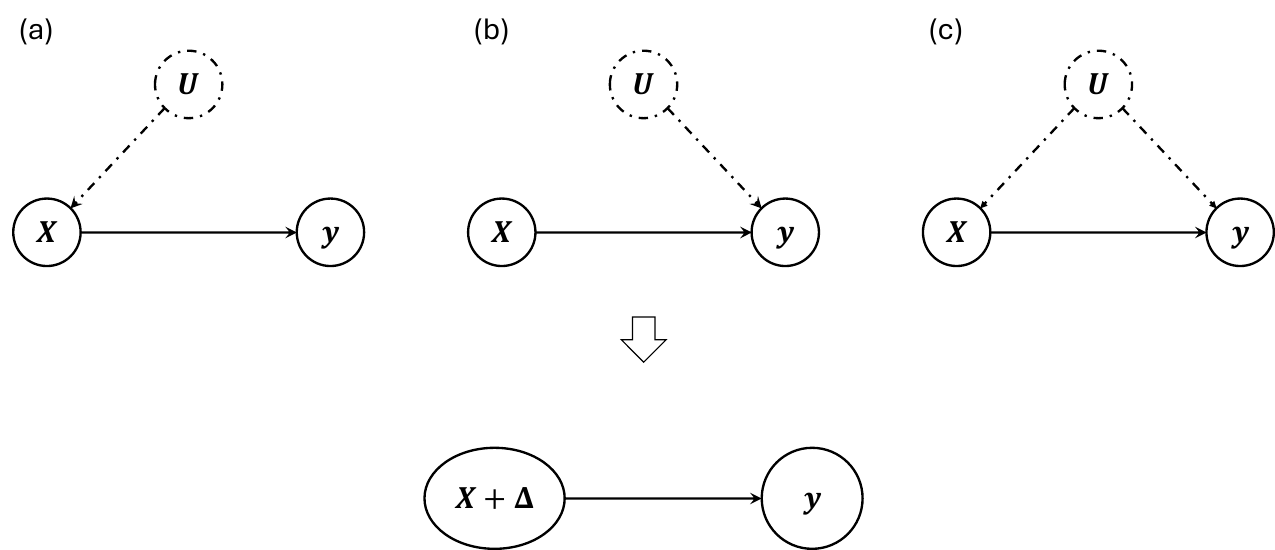}
    \caption{Three types of lurking variable impacts: (a) lurking variables $\boldsymbol{U}$ only alter the distributions of input $\boldsymbol{X}$, (b) only alter output $\boldsymbol{y}$, or (c) alter both, all of which can be captured by the equivalent transformation.}
    \label{fig: 1}
\end{figure}

The feasibility of feature-space adaptation is grounded in the idea of effect equivalence modeling (EEM) of lurking variables \cite{sabbaghi2018model}: given observed inputs and outputs, EEM compensates the influence of unobservable downstream factors by mapping their  effect onto the observed features within the column space of the pre-trained weights. The unobservable factors that cause domain discrepancy can be more formally described as lurking variables in statistics \cite{box1966use,joiner1981lurking}. A lurking variable is defined as a variable that has an important effect but is not included among the predictor variables under consideration. It may be omitted from the analysis because its existence is unknown, its influence is assumed to be negligible, or relevant data are unavailable \cite{hunter1979hazardous}. As illustrated in Fig.~\ref{fig: 1}, lurking variables $\boldsymbol{U}$ can alter the probability distributions of both input $\boldsymbol{X}$ and response $\boldsymbol{y}$ in the observed data, thus affecting the observed association between them. This leads to a perceived inconsistency between the downstream data and the pre-trained data. By identifying the equivalent amount $\boldsymbol{\Delta}$ of lurking variable effects through input transformation, the pre-trained model can be applied more effectively to the downstream domain without modifying its parameters.

Lurking variables are ubiquitous across application domains. In computer vision, style, texture, or lighting conditions can be viewed as lurking variables, systematically altering observed features while remaining unobserved and unlabeled during training \cite{mai2024fine}. In natural language processing, language type or domain (e.g., news versus reports) influences syntactic and lexical patterns, yet is often ignored in modeling \cite{brown2020language}. More generally, distributional discrepancies between pre-training datasets (e.g., ImageNet, Wikipedia) and downstream benchmarks (e.g., Oxford Pets, CoLA) can be viewed as arising from lurking variables that affect the input–output relationship in downstream domains or tasks. In manufacturing applications, unobservable process changes such as machine calibration have been effectively modeled as lurking variables in 3D printing quality control \cite{huang2015optimal,sabbaghi2018model}. In causal inference, unmeasured confounding remains a classic example of lurking variables, which has been extensively studied in synthetic control \cite{abadie2003economic}. 

The remainder of this paper is organized as follows. In Section 2, we formally review the notion of intruder dimensions, their link to forgetting, and the EEM framework for lurking variables, which together provide the foundation of our study. In Section 3, we show how EEM can be utilized to design a new feature-space fine-tuning strategy. In Section 4, we present a theoretical analysis that introduces a new forgetting measure for pre-trained models based on intruder dimensions and explains why our fine-tuned model can explicitly control forgetting. Section 5 reports empirical evaluations, demonstrating the effectiveness of the proposed approach, and Section 6 concludes the paper and outlines directions for future work.

\section{Preliminaries}

\subsection{Intruder Dimension}

\citet{shuttleworth2024lora} introduced the concept of \emph{intruder dimension} and discussed its impact on LoRA forgetting. In that work, the “direction” change between the pre-trained and fine-tuned weights was quantified by cosine similarity between the leading left singular vectors of the pre-trained and fine-tuned weight matrices obtained from their SVD decompositions. Although this is intuitive, cosine similarity is in general not a suitable notion of direction for weight matrices, because it is not {invariant} under natural symmetries of neural networks \cite{navon2023equivariant,theus2025generalized}. For example, permuting the hidden units (and hence permuting the rows/columns of a weight matrix) yields a permuted model that produces exactly the same outputs as the original pre-trained model for all pre-training inputs and is therefore functionally equivalent to it (see the detailed discussion in the appendix \ref{cs_nps}). Nevertheless, such a permutation will typically change the cosine similarity between the corresponding singular vectors, so a drop in cosine similarity in this case cannot be interpreted as the model having “forgotten” any knowledge. By contrast, the definition we introduce below depends only on the \emph{singular subspaces} of the pre-trained and fine-tuned weights, and is therefore \emph{invariant under such permutation symmetries}. 

Let $\boldsymbol{W}_0 = \boldsymbol{U}_0 \boldsymbol{\Sigma}_0 \boldsymbol{V}_0^\top$ be the singular value decomposition of the pre-trained weight matrix. Although $\boldsymbol{W}_0$ is typically full rank in a mathematical sense, it is not full rank numerically because its tail singular values are very small (e.g. $1\times10^{-5}$) (see the detailed discussion in the appendix \ref{rank_analysis_pretrained}). To make the spectral analysis meaningful, we define

\begin{equation}
    r_0 \;=\; \inf \Bigl\{ r \in \mathbb{Z}:\, \sum_{i = 1}^{r} \sigma_{0,i}^2 \,\geq\, \alpha \bigl\Vert \boldsymbol{W}_0 \bigr\Vert_F^2 \Bigr\},
    \label{rank: pre-trained}
\end{equation}
and, throughout this paper, approximate \(\boldsymbol{W}_0\) using only its leading \(r_0\) singular components as
\(    \boldsymbol{W}_0 \;\approx\;
    \boldsymbol{U}_0[:,1{:}r_0]\,
    \boldsymbol{\Sigma}_0[1{:}r_0,1{:}r_0]\,
    \boldsymbol{V}_0[:,1{:}r_0]^\top\). Let $\boldsymbol{W}_{\mathrm{ft}} = \boldsymbol{U}_{\mathrm{ft}} \boldsymbol{\Sigma}_{\mathrm{ft}} \boldsymbol{V}_{\mathrm{ft}}^\top$ be the singular value decomposition of the fine-tuned weight matrix, and let $r_{\mathrm{ft}}$ denote the number of its nonzero singular values. We define $\mathcal{S}_0 = \mathrm{span}\big(\boldsymbol{U}_0[:,1{:}r_0]\big)$, and let $\mathcal{S}_{\mathrm{ft}}$ be the subspace spanned by the left singular vectors of $\boldsymbol{W}_{\mathrm{ft}}$ corresponding to its nonzero singular values. A unit vector $\boldsymbol{u} \in \mathcal{S}_{\mathrm{ft}}$ is an \emph{intruder dimension} of $\boldsymbol{W}_{\mathrm{ft}}$ w.r.t. $\boldsymbol{W}_0$ if 
\(\|P_{\mathcal{S}_0} \boldsymbol{u}\|_2^2 < \varepsilon\),
where $P_{\mathcal{S}_0}$ is the orthogonal projector onto $\mathcal{S}_0$. 
In this paper, \emph{we will establish a mathematical link between intruder dimensions and catastrophic forgetting, and propose a principled approach to mitigating the impact of such intruder dimensions}.

\subsection{Effect Equivalance Modeling of Lurking Variables}

In modeling a system’s responses $\boldsymbol{y} \in \mathbb{R}^k$, it is useful to distinguish two categories of influencing variables: (i) observable variables $\boldsymbol{X}$ that are measured and available for analysis, and (ii) lurking variables $\boldsymbol{U}$ that are unobserved, ignored, or lack corresponding data yet may introduce variability \cite{box1966use,hunter1979hazardous}. Standard assumptions typically treat $\boldsymbol{X}$ as the sole explanatory factors and regard $\boldsymbol{U}$ as fixed or negligible. Under these assumptions, the system is modeled as
\(\boldsymbol{y} \;=\; f(\boldsymbol{X}) \;+\; \boldsymbol{\epsilon}\), where $f: \mathbb{R}^d \to \mathbb{R}^k$ denotes the response map and $\boldsymbol{\epsilon}$ is a $k$-dimensional noise term ($\mathbb{E}[\boldsymbol{\epsilon}] = \mathbf{0}$ and $\mathrm{Cov}(\boldsymbol{\epsilon}) = \boldsymbol{\Sigma}$). However, when such assumptions are violated (more aligned with real-world scenarios such as the downstream data with unknown discrepancy with the pre-training data), the response should instead be modeled as
{
\setlength{\abovedisplayskip}{2pt}
\setlength{\belowdisplayskip}{2pt}
\begin{equation}
 \boldsymbol{y} = g(\boldsymbol{X}, \boldsymbol{U}) + \boldsymbol{\epsilon}
 \label{eq: 1}
\end{equation}
}
where lurking variables \(\boldsymbol{U}\) exert non-negligible influence. When the usual assumptions are met,  Eq. \ref{eq: 1} is degenerated into Eq. \ref{eq: 2} and $f(\cdot)$ has a simpler form.
\begin{equation}
\boldsymbol{y} = g(\boldsymbol{X}, \boldsymbol{U} = \boldsymbol{0}) + \boldsymbol{\epsilon} = f(\boldsymbol{X}) + \boldsymbol{\epsilon}
\label{eq: 2}
\end{equation}

However, when lurking variables are present, they pose a challenging question: how can one infer and account for their effects when direct observation is not accessible?
The work in \cite{huang2015optimal,sabbaghi2018model} developed the EEM approach to model the effects of lurking variables to improve 3D printing accuracy. 

\textbf{Definition 1 (EEM).} 
Let $g : \mathbb{R}^d \times \mathbb{R}^m \rightarrow \mathbb{R}^k$ be a $k$-dimensional response function defined on observable variables $\boldsymbol{X}$ and lurking variables $\boldsymbol{U}$. 
Assume $g \in C^1(\mathbb{R}^d \times \mathbb{R}^m)$ and that for each $i \in \{1, \ldots, d\}$, the partial derivative $\frac{\partial g}{\partial x_i}(\boldsymbol{X}, \boldsymbol{0}) \in \mathbb{R}^k$
is non-zero. The total equivalent amount of lurking-variable effects $\boldsymbol{\Delta}$ in terms of $\boldsymbol{X}$ can then be estimated from the data.
\begin{equation}
g(\boldsymbol{X}, \boldsymbol{U})
= g(\boldsymbol{X}+\boldsymbol{\Delta}, \boldsymbol{U}=\boldsymbol{0})
= f(\boldsymbol{X}+\boldsymbol{\Delta})
\label{eq:eem_justification}
\end{equation}
The justification follows from the mean value theorem applied to a first-order Taylor expansion. There is no explicit solution for $\boldsymbol{\Delta}$, but it can be estimated through statistical learning or neural networks using the observed variables $\boldsymbol{X}$ as input. 

Lurking variables can explain the source of discrepancy between the pre-training dataset and the downstream dataset: unobserved or missing features prevent the pre-trained model from being directly applied to the downstream domain or task. EEM therefore provides a solution to mitigate the influence of lurking variables and also \emph{offers a perspective for performing fine-tuning in the feature space without changing model parameters}.

\section{Methodology}
\subsection{Problem Setup}

We evaluate the robustness and forgetting of fine-tuning methods using established criteria from the literature \cite{wortsman2022robust}: fine-tuning on a single downstream dataset should not reduce accuracy on other downstream datasets compared to the zero-shot model. In this work, robustness is interpreted as inversely related to forgetting: more robust methods are viewed as inducing less forgetting.

Firstly, we present the general formulation for model fine-tuning. Let $f_{\boldsymbol{W}_0}:\mathcal{X}\!\to\!\mathcal{Y}$ be a pre-trained model with parameters $\boldsymbol{W}_0$. Given a downstream dataset $\mathcal{D}_{\mathrm{ft}}$, fine-tuning solves
{
\setlength{\abovedisplayskip}{2pt}
\setlength{\belowdisplayskip}{2pt}
\begin{equation*}
\begin{aligned}
\boldsymbol{W}_{\mathrm{ft}} 
&= \arg\min_{\boldsymbol{W} \in \mathcal{A}} 
\mathcal{L}(\boldsymbol{W};\mathcal{D}_{\mathrm{ft}}),\\
\mathcal{L}(\boldsymbol{W};\mathcal{D}_{\mathrm{ft}})
&=\mathbb{E}_{(\boldsymbol{X},\boldsymbol{y})\sim\mathcal{D}_{\mathrm{ft}}}
\big[\ell(\boldsymbol{X},\boldsymbol{y};\,f_{\boldsymbol{W}})\big],
\end{aligned}
\end{equation*}

where $\mathcal{A}$ encodes the adaptation family (e.g., full fine-tuning and LoRA) and $\mathcal{L}(\boldsymbol{W};\mathcal{D}_{\mathrm{tr}})$ denotes the loss function of $\boldsymbol{W}$ for fine-tuning dataset $\mathcal{D}_{\mathrm{tr}}$. 

In practice, the choice of loss function $\mathcal{L}$ depends on the downstream task: cross-entropy for classification, mean squared error for regression, and token-level negative log-likelihood for natural language generation tasks.

For full fine-tuning, $\boldsymbol{W}$ is initialized at the pre-trained parameters $\boldsymbol{W}_0 = 
\{\boldsymbol{W}_0^{(l)}\}_{l=1,...,L}$ (where $L$ denotes the number of learnable layers in the pre-trained model), and update all components end to end. For LoRA, each layer weight $\boldsymbol{W}_0^{(\ell)} \in \mathbb{R}^{p\times q}$ is frozen and augmented with a low-rank update $\mathbf{B}^{(\ell)}\mathbf{A}^{(\ell)}$, where $\mathbf{A}^{(\ell)} \in \mathbb{R}^{r\times q}$ and $\mathbf{B}^{(\ell)} \in \mathbb{R}^{p\times r}$ with $r \ll \min(p,q)$. The fine-tuned weight is 
{
\setlength{\abovedisplayskip}{2pt}
\setlength{\belowdisplayskip}{2pt}
\begin{equation}
    \boldsymbol{W}_{\mathrm{ft}}^{(\ell)} = \boldsymbol{W}_0^{(\ell)} + \mathbf{B}^{(\ell)}\mathbf{A}^{(\ell)}
    \label{lora}
\end{equation}
}
and only the low-rank parameters $\mathbf{A}^{(\ell)},\mathbf{B}^{(\ell)}$ are optimized.

Secondly, we consider cross-dataset robustness in the proposed fine-tuning formulation. Under this setup, we assess how fine-tuning on one downstream domain/task affects zero-shot generalization to a different domain/task under distribution shift. Concretely, suppose we fine-tune the pre-trained model $f_{\boldsymbol{W}_0}$ on a dataset $\mathcal{D}^{(a)}_{\mathrm{ft}}$, obtaining the adapted parameters $\boldsymbol{W}_{\mathrm{ft}}$. The resulting model $f_{\boldsymbol{W}_{\mathrm{ft}}}$ is assessed not only on the in-domain/task test set $\mathcal{D}^{(a)}_{\mathrm{te}}$, but also on out-of-domain/task test sets $\{\mathcal{D}^{(b)}_{\mathrm{te}} : b \neq a\}$ corresponding to other downstream datasets. {Robustness} is measured by the relative change in task-specific metric $\mathcal{M}$ (where larger values indicate better performance) compared to the zero-shot pre-trained model $f_{\boldsymbol{W}_0}$:
\begin{equation}
R^{(b)} = \mathcal{M}(f_{\boldsymbol{W}_{\mathrm{ft}}}, \mathcal{D}^{(b)}_{\mathrm{te}})
          - \mathcal{M}(f_{\boldsymbol{W}_0}, \mathcal{D}^{(b)}_{\mathrm{te}}).
\end{equation} 
A fine-tuned model $f_{\boldsymbol{W}_{\mathrm{ft}}}$ is \emph{robust} if $R^{(b)} \ge -\epsilon$ for all $b \neq a$.

\subsection{Vector-based Feature Space Adaptation (VeFA)}

Rather than blindly updating all parameters or indiscriminately altering the pre-trained model’s weight space, one should first understand how the downstream distribution differs from the pre-trained dataset. For example, in image classification, class semantics are largely stable across domains, while discrepancies typically arise from style, background, resolution, illumination, viewpoint, and other contextual factors. These factors act as lurking variables that confound the input while leaving intrinsic class semantics unchanged.

Inspired by the EEM view of lurking variables, we propose to perform fine-tuning directly in the feature space via one simple instantiation, Vector-based Feature Space Adaptation (VeFA). Concretely, we keep $\boldsymbol{W}_0^{(\ell)}$ frozen and introduce a diagonal feature-wise scaling matrix $\boldsymbol{\Lambda}^{(\ell)}$, yielding the following model for the $\ell$-th layer:
\begin{equation*}
\begin{aligned}
\boldsymbol{h}^{(\ell)}&= \boldsymbol{W}_0^{(\ell)}\left( \boldsymbol{h}^{(\ell-1)}+\boldsymbol{\lambda}^{(\ell)}\odot \boldsymbol{h}^{(\ell-1)} \right)\\
&=\boldsymbol{W}_0^{(\ell)}\!\left( \boldsymbol{I} + \boldsymbol{\Lambda}^{(\ell)} \right)\boldsymbol{h}^{(\ell-1)}=\boldsymbol{W}_{\mathrm{ft}}^{(\ell)} \boldsymbol{h}^{(\ell-1)}
\end{aligned}
\end{equation*}
where $\boldsymbol{I}$ is the identity matrix and $\boldsymbol{\Lambda}^{(\ell)}=diag \left( \boldsymbol{\lambda}^{(\ell)} \right)$ performs channel-wise scaling of the layer input. Essentially, compared to LoRA (Fig. \ref{fig: schematic comparistion}a), the updating rule of VeFA (Fig. \ref{fig: schematic comparistion}b) is:
\begin{equation}
    \boldsymbol{W}_{\mathrm{ft}}^{(\ell)} = \boldsymbol{W}_0^{(\ell)} + \boldsymbol{W}_0^{(\ell)}\boldsymbol{\Lambda}^{(\ell)}
    \label{lora}
\end{equation}
The vector $\boldsymbol{\lambda}$ is initialized to zero. 

Compared to LoRA, VeFA is substantially more parameter-efficient. To our knowledge, VeFA actually ranks among the most parameter-efficient LoRA-style variants, especially compared to approaches such as VeRA \cite{kopiczko2023vera}. In addition, all VeFA updates are constrained to the column space of the pre-trained weight matrix: $\mathrm{Col}(\boldsymbol{W}_0 \boldsymbol{\Lambda}^{(\ell)}) \;\subseteq\; \mathrm{Col}(\boldsymbol{W}_0)$. VeFA preserves the representation subspace of $\boldsymbol{W}_0$ and does not introduce any intruder dimensions, as illustrated in
Fig. \ref{fig: schematic comparistion}(c).

\begin{figure*}[htbp]
    \centering
    \includegraphics[width=0.7\textwidth]{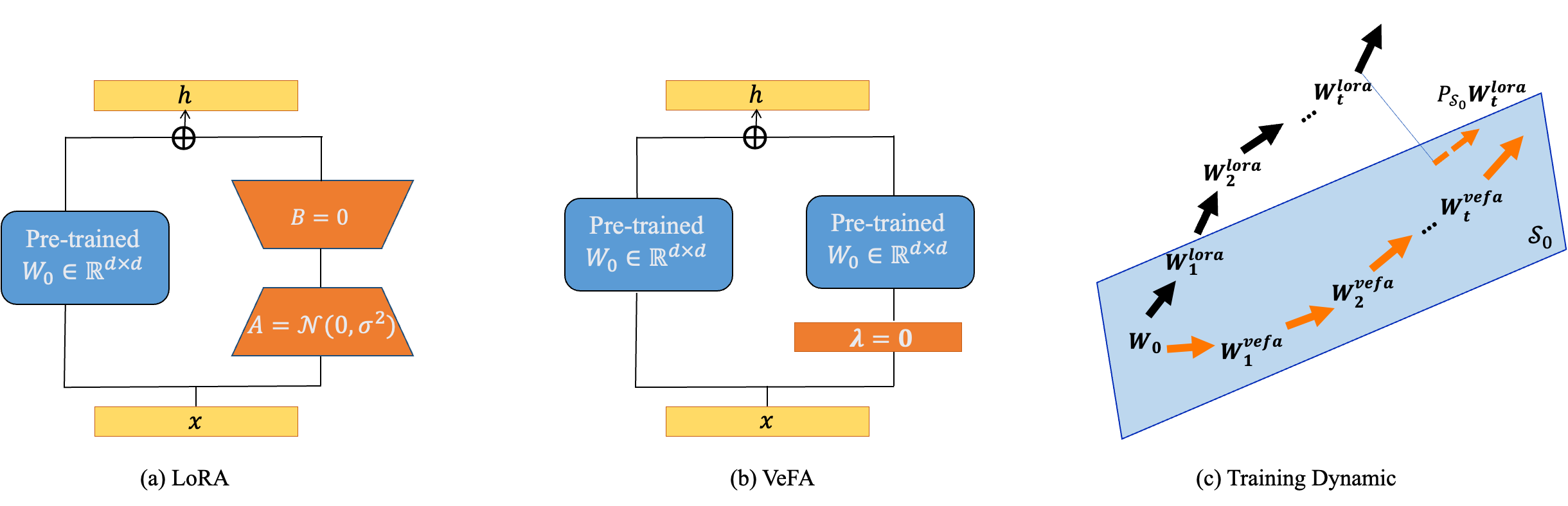}
    \caption{Schematic comparison of LoRA (a), VeFA (b) and their training dynamics (c). LoRA updates the weights matrix $\boldsymbol{W}$ by training the low-rank matrices $\boldsymbol{A}$ and $\boldsymbol{B}$, with intermediate rank $r$. VeFA keeps the pre-trained $\boldsymbol{W}_0$ frozen and performs adaptation at the element-wise feature level.}
    \label{fig: schematic comparistion}
\end{figure*}

We restrict adaptation to channel-wise scaling of the hidden representation without cross-channel mixing for practical reasons. First, recent work on \emph{sparsifying large neural networks} (e.g., sparse attention post-training for mechanistic interpretability \cite{draye2025sparse} and weight-sparse transformers with interpretable circuits \cite{gao2025weight}) suggests that modern large-scale models are highly over-parameterized and contain substantial redundancy: many weights can be pruned or sparsified with little loss in performance. Second, {downstream fine-tuning datasets are typically much smaller and semantically narrower than the pre-training datasets}, implying that the effective function required for the downstream task is often considerably simpler than the original pre-trained model. Therefore, the information encoded in the pre-trained model is even more redundant. 

To compare weight-space and feature-space adaptation schemes, we first give a conceptual illustration. For weight-space adaptation, fine-tuning treats the neurons as fixed and updates the connection weights between them. Both hidden layers and the output layer acquire new trainable parameters. For feature-space adaptation, the original connection weights are frozen, and only lightweight, neuron-wise transformations are learned on top of the pre-trained features. This keeps the representational backbone intact, while still allowing task-specific adaptation through feature scaling. We also provide one numerical example on Appendix \ref{numerical_example_weight_feature}.

It is worth noting that our method has a different focus from existing feature adaptation research. In prior work, “features” refer to the outputs of a feature extractor, which are then modified to improve downstream transfer—such as CLIP adaptation \cite{gao2024clip}, incremental learning \cite{iscen2020memory}, or domain adaptation \cite{chen2019progressive,zheng2020cross}.  In contrast, we perform layer-wise feature adaptation to avoid introducing intruder dimensions.

\section{Theoretical Analysis}

In this section, we introduce a mathematical interpretation of how forgetting is related to \emph{intruder dimensions}, and rigorously analyze why VeFA can effectively control forgetting based on the interpretation. We focus on a linear regression setting with mean-squared loss. The data distribution of the downstream task is generated by the underlying true model
\(\boldsymbol{y} = \boldsymbol{W}^* \boldsymbol{x} + \boldsymbol{\varepsilon}\) with i.i.d.\ noise \(\boldsymbol{\varepsilon}\). Let \(\boldsymbol{W}_0\) and
\(\boldsymbol{W}_{\mathrm{ft}} = \boldsymbol{W}_0 + \boldsymbol{\Delta}\) denote the pre-trained and fine-tuned weight matrices, respectively. The rank of $\boldsymbol{W}_0$ is $r_0$, as defined in Eq. \ref{rank: pre-trained}.

Under this setting, the increase in the expected loss over the data distribution (expected risk) is studied:
\begin{equation}
\mathcal{L} (\boldsymbol{W}_{\mathrm{ft}} \vert \boldsymbol{x}, \boldsymbol{y}) - \mathcal{L} (\boldsymbol{W}_0 \vert \boldsymbol{x}, \boldsymbol{y}) = \text{Tr} \left( \boldsymbol{\Delta} \boldsymbol{\Sigma}_{\boldsymbol{x}} \boldsymbol{\Delta}^T \right)
\label{forgetting: overall}
\end{equation}
where $\boldsymbol{x}$ and $\boldsymbol{y}$ follow the pre-training data distribution, and $\Sigma_{\boldsymbol{x}} = \mathbb{E}(\boldsymbol{x}\boldsymbol{x}^{\top})$ denotes the empirical covariance matrix of the pre-training data. 

This increase in the expected loss quantifies the degradation in model generalization after fine-tuning, capturing effects related to catastrophic forgetting and reduced robustness. A straightforward approach implied by Eq. \ref{forgetting: overall} is to regulate the norm of $\boldsymbol{\Delta}$ (that is, to impose a constraint on the magnitude of the update), which is exactly the key strategy adopted by many works on mitigating catastrophic forgetting \cite{kirkpatrick2017overcoming} and on robust fine-tuning\cite{gouk2020distance,tian2023fast}.

In this paper, we carry out an in-depth analysis of the increased expected loss  by decomposing Eq. \ref{forgetting: overall} and  linking it to the intruder dimension. Recall that $\mathcal{S}_0$ denotes the column space of $\boldsymbol{W}_0$, let \(P_{\mathcal{S}_0}\) and \(P_{\mathcal{S}_0}^{\perp}\) be the orthogonal projections onto \(\mathcal{S}_0\) and its orthogonal complement. Then the update \(\boldsymbol{\Delta}\) admits the decomposition as $\boldsymbol{\Delta} = P_{\mathcal{S}_0} \boldsymbol{\Delta} + P_{\mathcal{S}_0}^\perp \boldsymbol{\Delta}=\boldsymbol{\Delta}_\parallel +
\boldsymbol{\Delta}_\perp$. Then
\begin{equation}
\label{forgetting: decomposition}
\text{Tr}\bigl(\boldsymbol{\Delta}\boldsymbol{\Sigma}_{\boldsymbol{x}}\boldsymbol{\Delta}^\top\bigr)
= \text{Tr} \bigl(\boldsymbol{\Delta}_\parallel \boldsymbol{\Sigma}_{\boldsymbol{x}}
            \boldsymbol{\Delta}_\parallel^\top\bigr)
+ \text{Tr} \bigl(\boldsymbol{\Delta}_\perp \boldsymbol{\Sigma}_{\boldsymbol{x}}
            \boldsymbol{\Delta}_\perp^\top\bigr).
\end{equation}
The cross term is canceled as the result of cyclic property of trace and $P_{\mathcal{S}_0} P_{\mathcal{S}_0}^\perp = \boldsymbol{0}$. We can see that both terms in the decomposition (\ref{forgetting: decomposition}) are non-negative, and the positive loss increase is expected since $\boldsymbol{W}_0$ is the best estimator for the pre-training data. To the best of our knowledge, Eq. \ref{forgetting: overall} and \ref{forgetting: decomposition} provide the first general formulation to quantify forgetting in terms of the increase in expected risk, independent of specific fine-tuning mechanism such as LoRA or its variants}. This quantitative metric is not restricted to evaluating PEFT approaches; in principle, it can be used in a wide variety of forgetting-related scenarios, including continual learning.

The first loss term of in Eq. \ref{forgetting: decomposition} corresponds to the effect of transformations within the column space of $\boldsymbol{W}_0$, while the second term captures the forgetting or harmful effect introduced by intruder dimensions. For VeFA, the second loss term in Eq. \ref{forgetting: decomposition} is zero, which means that  VeFA does not suffer from the intruder-dimension–induced forgetting that arises in LoRA. 

Eq. \ref{forgetting: decomposition} also suggests that when intruder dimensions are introduced, this second loss term becomes strictly positive. To quantify the additional forgetting purely attributed to intruder directions, Proposition \ref{intruder_dimension_loss} below provides an explicit lower bound of the increased loss. The proof of Proposition \ref{intruder_dimension_loss} can be found in Appendix \ref{proof_intruder_dimension_loss}.

\begin{proposition}
\label{intruder_dimension_loss}
If $P_{\mathcal{S}_0}^\perp \boldsymbol{\Delta} \neq \boldsymbol{0}$, the increase in expected loss attributable to the intruder dimensions admits a lower bound:
{
\setlength{\abovedisplayskip}{2pt}
\setlength{\belowdisplayskip}{2pt}
\begin{equation*}
\label{eq:lower-bound}
\begin{aligned}
\text{Tr}\bigl(\boldsymbol{\Delta}_\perp \boldsymbol{\Sigma}_{\boldsymbol{x}} \boldsymbol{\Delta}_\perp^\top\bigr)
&\;\ge\;
\sigma_{\min}\bigl(\boldsymbol{\Sigma}_{\boldsymbol{x}}\bigr)
\Bigl(
  \sqrt{(1-\varepsilon)\, r_{\mathrm{ft}}\, \sigma_{{\mathrm{ft}},r_{\mathrm{ft}}}^2}
\\[-0.25em]
&\qquad\qquad
  - \sqrt{1-\alpha}\,\|\boldsymbol{W}_0\|_F
\Bigr)_+^2.
\end{aligned}
\end{equation*}
}
\end{proposition}

For VeFA, we further prove that the first loss term in Eq. \ref{forgetting: decomposition} has an upper bound, as specified by Proposition \ref{in_subspace_loss} (proof can be seen in Appendix \ref{proof_in_subspace_loss}). Notably, the bound depends only on the maximum absolute entry of the diagonal vector $\boldsymbol{\lambda}$, so enforcing $\|\boldsymbol{\lambda}\|_\infty \le \gamma$ yields a global bound on the total forgetting under VeFA. In the next section, we also show with experiments that, for most tasks, $\max_j |\lambda_j|$ of the fine-tuned matrices remains small even without any explicit constraint. For more challenging tasks, one can additionally impose an explicit upper bound on $\max_j |\lambda_j|$, which directly controls the overall forgetting (seen in Appendix \ref{appendix: Maximum VeFA Scaling Coefficient}).
\begin{proposition}
\label{in_subspace_loss}
Assume that $\boldsymbol{\Sigma}_{\boldsymbol{x}}$ and $\|\boldsymbol{W_0}\|_F$ are bounded. For VeFA, the in-subspace loss term satisfies 
{
\setlength{\abovedisplayskip}{2pt}
\setlength{\belowdisplayskip}{2pt}
\[
\text{Tr} \left( P_{\mathcal{S}_0} \boldsymbol{\Delta} \boldsymbol{\Sigma}_{\boldsymbol{x}} \boldsymbol{\Delta}^T P_{\mathcal{S}_0}^T \right) \le\;
\|\boldsymbol{\Sigma}_{\boldsymbol{x}}\|_2 \,\|\boldsymbol{W}_0\|_F^2
\bigl(\max_j |\lambda_j|\bigr)^2 .
\]
}
\end{proposition}

In summary, VeFA introduces no intruder dimensions and, therefore, avoids the associated intruder-induced forgetting. Within the original main column space of $\boldsymbol{W}_0$, the transformation loss is tightly bounded and controlled by the small feature-wise scaling factors $\boldsymbol{\lambda}$, so the overall forgetting under VeFA remains bounded. Consequently, VeFA exhibits substantially reduced catastrophic forgetting and, correspondingly, strong robustness.

\section{Experiment}
\subsection{Experiment setup}

\textbf{Pre-trained models and downstream tasks}. To demonstrate the effectiveness of VeFA, we fine-tune three different pre-trained models on their corresponding downstream datasets. Our goals are twofold: (i) to show that VeFA achieves downstream performance comparable to LoRA while using substantially fewer trainable parameters, and (ii) to examine, through multiple cross-task evaluations, whether VeFA attains better performance than LoRA in terms of reduced forgetting, and thus improved robustness. 

1. Image Classification (CLIP across seven datasets \cite{radford2021learning,zanella2024low}). We fine-tune CLIP (ViT-B/16 backbone) on seven image classification datasets. In each case, the model is fine-tuned on one dataset and then evaluated on the others. By comparing fine-tuned performance with CLIP’s zero-shot baseline, we assess the cross-dataset robustness of different fine-tuning strategies.

2. Natural Language Understanding (RoBERTa on GLUE \cite{liu2019roberta,wang2018glue}). We evaluate RoBERTa on the GLUE benchmark. Since GLUE uses task-specific classification heads, we consider two complementary evaluation settings for cross-task robustness. (1) Pairwise cross-task evaluation. For each source task, we fine-tune an adapter on its training set and evaluate the resulting model on the remaining GLUE tasks. (2) Merge-based cross-task evaluation. For each target task, we keep its fine-tuned head fixed and merge adapters trained on the other tasks into a shared backbone. By default, we use federated averaging (FedAvg) \cite{mcmahan2017communication}. Since GLUE tasks may exhibit non-trivial correlations, we further stress-test the merging procedure by sampling the merging coefficients from a Dirichlet distribution with hyperparameter $\tau$ and forming a convex combination of adapters \cite{chen2024pareto}. The resulting performance, compared to the zero-shot RoBERTa baseline on the same task, serves as our robustness measure for NLU.

3. Natural Language Generation (GPT-2 on E2E \cite{radford2019language,novikova2017e2e}). We fine-tune GPT-2 on the E2E NLG benchmark, which provides a relatively data-rich downstream setting compared to typical few-shot settings. Here, the goal is to assess effectiveness: we test whether VeFA’s constrained, channel-wise adaptation can still match LoRA’s performance when more downstream data are available.

\subsection{Fine-tuning CLIP for Image Classification}

We follow the setting of previous work \cite{zhou2022learning,zanella2024low}. We evaluated seven datasets spanning diverse visual domains—satellite imagery (EuroSAT \cite{helber2019eurosat}), food (Food101 \cite{bossard2014food}), pet breeds (OxfordPets \cite{parkhi2012cats}), flowers (Flower102 \cite{nilsback2008automated}), generic objects (Caltech101 \cite{fei2004learning}), textures (DTD \cite{cimpoi2014describing}), and human actions (UCF101 \cite{soomro2012ucf101}). Together, these datasets provide a comprehensive benchmark for visual classification.

We compare VeFA with prompt-based methods, adapter-based methods, and LoRA (rank $r=2$) in terms of few-shot performance, using a total of eight baselines. We use $r=2$ to match the setting in previous work \cite{zanella2024low}, where it is shown to be sufficient for a few-shot adaptation. In few-shot learning, a shot refers to the number
of labeled training examples provided per class. We conducted experiments under 1-shot, 4-shot, and 16-shot learning settings. The results for VeFA and LoRA are reported in Tab.~\ref{Tab: few-shot-image}. A complete set of results on effectiveness and robustness is provided in Appendices \ref{Complete effectivess_visual} and \ref{Complete robustness_visual}, respectively. For the 1-shot and 4-shot settings, we report only effectiveness, whereas for the 16-shot setting, we report both effectiveness and robustness (see the last two rows of Tab. \ref{Tab: few-shot-image}). With only 25\% of LoRA’s trainable parameters, VeFA achieves higher average performance in the 1-shot, 4-shot, and 16-shot settings and consistently demonstrates stronger robustness across all seven datasets. In addition, VeFA achieves the best accuracy among all competing baselines. These results highlight VeFA’s advantage in low-data settings, enabling effective knowledge transfer while better preserving pre-trained representations.

\begin{table*}
\caption{Performance comparison between LoRA and VeFA across different datasets under zero-shot, 1-shot, 4-shot,
16-shot fine-tuning, and robustness. The visual backbone is ViT-B/16. Better results are in \textbf{bold}.}
\label{Tab: few-shot-image}
\centering

\setlength{\tabcolsep}{5pt}
\renewcommand{\arraystretch}{0.8}

\begin{footnotesize}

\begin{tabular}{c|c|c|c|c|c|c|c|c}
\toprule
Shots & \textbf{Type} & \textbf{Caltech101} & \textbf{Food101} & \textbf{Oxford Pets} &
\textbf{Oxford Flowers} & \textbf{EuroSAT} & \textbf{DTD} & \textbf{UCF101} \\
\midrule
0  & CLIP & 93.2 & 85.2 & 88.2 & 67.3 & 42.3 & 44.3 & 65.0 \\
\midrule
\multirow{2}{*}{1}
 & LoRA & 94.0 & 82.3 & 91.8 & 79.4 & 71.8 & 53.0 & \textbf{75.4} \\
 & VeFA & \textbf{94.1} & \textbf{86.3} & \textbf{93.3} & \textbf{84.3} & \textbf{73.3} & \textbf{55.0} & 74.2 \\
\midrule
\multirow{2}{*}{4}
 & LoRA & 95.2 & 81.6 & 90.8 & \textbf{93.8} & \textbf{87.5} & 62.2 & 78.5 \\
 & VeFA & \textbf{95.6} & \textbf{86.0} & \textbf{93.4} & 93.0 & 87.1 & \textbf{65.7} & \textbf{80.2} \\
\midrule
\multirow{2}{*}{16}
 & LoRA  & 96.3 & 84.0 & 91.3 & \textbf{98.2} & 90.2 & 71.8 & 86.2 \\
 & VeFA  & \textbf{96.4} & \textbf{87.8} & \textbf{94.4} & 97.5 & \textbf{91.3} & \textbf{72.5} & \textbf{86.4} \\
\midrule
\multirow{2}{*}{$\bar{R}$}
 & LoRA & -2.4 & -3.8 & -2.8 & -4.4 & -14.1 & -3.0 & -1.1 \\
 & VeFA & \textbf{+0.2} & \textbf{-0.6} & \textbf{-0.4} & \textbf{+0.4} & \textbf{-2.3} & \textbf{+2.0} & \textbf{+0.3} \\
\bottomrule
\end{tabular}
\end{footnotesize}
\end{table*}

\subsection{Natural Language Understanding}

We evaluated on the General Language Understanding Evaluation (GLUE) benchmark \cite{wang2018glue} using RoBERTa-base and RoBERTa-large \cite{liu2019roberta}. We compare VeFA against LoRA and six other PEFT methods. Our experiment broadly follows the setting in LoRA \cite{hu2022lora}: we adapt each self-attention block and fully train the task-specific classification head. Unlike \cite{hu2022lora}, which employs an auxiliary hyperparameter $\alpha$ to rescale gradients in adapted layers, we use separate learning rates for (i) the classification head and (ii) the adapted layers. Learning rates and training epochs are chosen via hyperparameter tuning; full settings appear in Appendix \ref{appendix: hyperparameters}. We use batch size 64 for RoBERTa-base and 32 for RoBERTa-large.

In line with \citet{hu2022lora}, we report the number of trainable parameters in the backbone layers, explicitly excluding the task-specific classification head. For each setting, we run five trials with different random seeds, take the best epoch in each trial, and report the median over these five results. We also reproduce the original LoRA results under the same experimental setup as in \citet{hu2022lora} and observe that our SVD-initialized models achieve nearly identical performance, indicating that the SVD approximation does not affect fine-tuning. To evaluate cross-task robustness, we use the checkpoint with the median validation performance among the five runs. For model merging, we consider both federated averaging and Dirichlet-based merging with $\tau=10$ (near-uniform) and $\tau=1$ (diverse weighting). For each setting, we draw five independent samples of the merging coefficients.

The key results are summarized in Tab.~\ref{Tab: GLUE}. For completeness, Appendix~\ref{Complete effectivess_GLUE} reports all effectiveness results, and Appendix~\ref{Complete robustness_GLUE} reports all robustness results, including pairwise cross-task evaluation and different merging strategies. In terms of effectiveness, VeFA achieves performance comparable to LoRA and other PEFT methods across both backbones while using substantially fewer trainable parameters. Moreover, across all cross-task robustness settings, VeFA-based adapters show better compatibility and consistently outperform LoRA. These results indicate that VeFA provides stronger cross-task robustness than LoRA.

\begin{table*}[!t]
\caption{Accuracy and Robustness on the GLUE Benchmark. We report Matthews correlation for CoLA, Pearson correlation for STS-B, and accuracy for all other tasks; higher is better.
Results for all methods except VeFA and LoRA are taken from prior work
\cite{hu2022lora,kopiczko2023vera,zhang2023lora}.}
\label{Tab: GLUE}
\centering

\setlength{\tabcolsep}{5pt}
\renewcommand{\arraystretch}{1.0}

\begin{footnotesize}

\begin{tabular}{c|c|c|c|c|c|c|c|c|c|c}
\toprule
& & \textbf{Type} & \textbf{\# Params} & \textbf{SST-2} & \textbf{MRPC} & \textbf{CoLA} &
   \textbf{QNLI} & \textbf{RTE} & \textbf{STS-B} & \textbf{AVG} \\
\midrule
\multirow{10}{*}{\rotatebox{90}{\textbf{Acc}}} & \multirow{5}{*}{\textbf{B}}
 & FT     & 125M    & 94.8 & 90.2 & 63.6      & 92.8 & 78.7 & 91.2 & 85.2 \\
 & & LoRA   & 0.3M    & 94.7$\pm$0.2 & 90.1$\pm$0.9 & 64.7$\pm$1.6 & 92.6$\pm$0.2 &  86.6$\pm$0.9 & 91.4$\pm$0.1 & \textbf{86.7} \\
  & & LoRA-FA & 1.8M & 94.8& 90.0& 63.6& 92.5& 67.9& 89.6& 83.1 \\
  && VeRA   & 0.043M  & 94.6$\pm$0.1 & 89.5$\pm$0.5 & 65.6$\pm$0.8 & 91.8$\pm$0.2 &
              78.7$\pm$0.7 & 90.7$\pm$0.2 & 85.2 \\
  && VeFA (Ours)   & \textbf{0.018M} & 94.1$\pm$0.1 & 88.5$\pm$0.7 & 63.3$\pm$1.0 &
              91.7$\pm$0.2 & 83.0$\pm$0.4 & 90.7$\pm$0.1 & 85.2 \\
\cmidrule(lr){2-11}
 &\multirow{5}{*}{\textbf{L}}
 & FT & 355M & 96.4& 90.9& 68.0& 94.7& 86.6& 92.4& 88.1\\
  && LoRA   & 0.8M    & 96.2$\pm$0.2 & 89.7$\pm$1.1 & 66.0$\pm$2.0 &
              94.8$\pm$0.3 & 90.2$\pm$0.6 & 92.4$\pm$0.3 & \textbf{88.2} \\
 && LoRA-FA & 3.7M& 96.0& 90.0 & 68.0& 94.4& 86.1& 92.0& 87.7\\
 & & VeRA   & 0.061M  & 96.1$\pm$0.1 & 90.9$\pm$0.7 & 68.0$\pm$0.8 &
              94.4$\pm$0.2 & 85.9$\pm$0.7 & 91.7$\pm$0.8 & 87.8 \\
  && VeFA (Ours)  & \textbf{0.049M} & 95.8$\pm$0.2 & 90.4$\pm$0.8 & 68.0$\pm$1.2 &
              94.1$\pm$0.3 & 87.0$\pm$0.4 & 91.2$\pm$0.4 & 87.8 \\
\midrule
\multirow{4}{*}{\rotatebox{90}{\textbf{Rob}}} & \multirow{2}{*}{\textbf{B}} & LoRA ($\tau=1$) & 0.3 M & 15.4$\pm$0.9 & 0& 0.1$\pm$4.8& 2.3$\pm$1.5& 0& 6.1$\pm$7.1& 4.0\\
& & VeFA ($\tau=1$) & 0.018 M & 4.8$\pm$1.4 & 10.1$\pm$8.6 & 37.5$\pm$8.6  & 2.8$\pm$1.3 & 12.6$\pm$4.5& 44.9$\pm$3.9& \textbf{18.8}\\
\cmidrule(lr){2-11}
&  \multirow{2}{*}{\textbf{L}} & LoRA ($\tau=1$) & 0.3 M & -1.2$\pm$1.4& 0& 0& -4.6$\pm$1.3& -0.6$\pm$0.6& 15.4$\pm$9.1 & 1.5 \\
& &  VeFA ($\tau=1$) & 0.018 M & -0.6$\pm$3.1& 19.6$\pm$12.4 & 6.6$\pm$1.7 & 15.8$\pm$2.5 & 18.2$\pm$2.0 & 60.4$\pm$4.0 & \textbf{20.0} \\
\bottomrule
\end{tabular}
\end{footnotesize}
\end{table*}

\subsection{Natural Language Generation}

We evaluated on the E2E benchmark \cite{novikova2017e2e}, following the experimental protocol used for LoRA \cite{hu2022lora}, and fine-tuned GPT-2 Medium and Large \cite{radford2019language}. In this setting, the downstream task is more complex and the dataset is substantially larger than in our visual classification and NLU experiments. Our goal is to test whether VeFA still retains sufficient expressive power under such a challenging regime. For a fair comparison, we run LoRA and VeFA within the same experimental platform. For LoRA, we use the official hyperparameters from \cite{hu2022lora}, while a complete list of the hyperparameters used for VeFA is provided in Appendix~\ref{appendix: hyperparameters}. Moreover, we bound VeFA’s scaling vector by clipping each entry of $\boldsymbol{\lambda}$ to $[-2,2]$, so that $\|\boldsymbol{\lambda}\|_\infty \le 2$ throughout training.

The E2E benchmark contains a single task: given a meaning representation, generate natural-language descriptions. We evaluate with five metrics—BLEU, NIST, MET, ROUGE-L, and CIDEr—to comprehensively assess generation quality. We report the results from the final epoch. As shown in Tab.~\ref{Tab: E2E}, VeFA achieves performance comparable to LoRA on both GPT-2 Medium and GPT-2 Large, while using substantially fewer trainable parameters.

\begin{table}[t]
\centering
\caption{Performance comparison of LoRA and VeFA on GPT-2 Medium and GPT-2 Large using standard NLG evaluation metrics.}
\label{Tab: E2E}
\setlength{\tabcolsep}{2pt}
\renewcommand{\arraystretch}{1.2}

\begin{footnotesize}
\begin{tabular}{c|c|c|c|c|c|c|c}
\toprule
 & \textbf{Type} & \textbf{\# Params} & \textbf{BLEU} & \textbf{NIST} &
   \textbf{MET} & \textbf{ROUGE-L} & \textbf{CIDEr} \\
\midrule
\multirow{2}{*}{\rotatebox{90}{\textbf{Med}}}
 & LoRA & 0.35M & 67.04 & 8.5753 & 45.92 & 68.74 & 2.3507 \\
 & VeFA & 0.073M & 66.35 & 8.5628 & 44.79 & 66.93 & 2.2274 \\
\midrule
\multirow{2}{*}{\rotatebox{90}{\textbf{Large}}}
 & LoRA & 0.77M & 67.38 & 8.6293 & 45.98 & 68.82 & 2.3320 \\
 & VeFA & 0.14M & 67.38 & 8.5889 & 46.11 & 68.86 & 2.3684 \\
\bottomrule
\end{tabular}
\end{footnotesize}
\end{table}

\subsection{Ablation Study and Further Analysis}

\begin{figure}[htbp]
    \centering
    \includegraphics[width=0.36\textwidth]{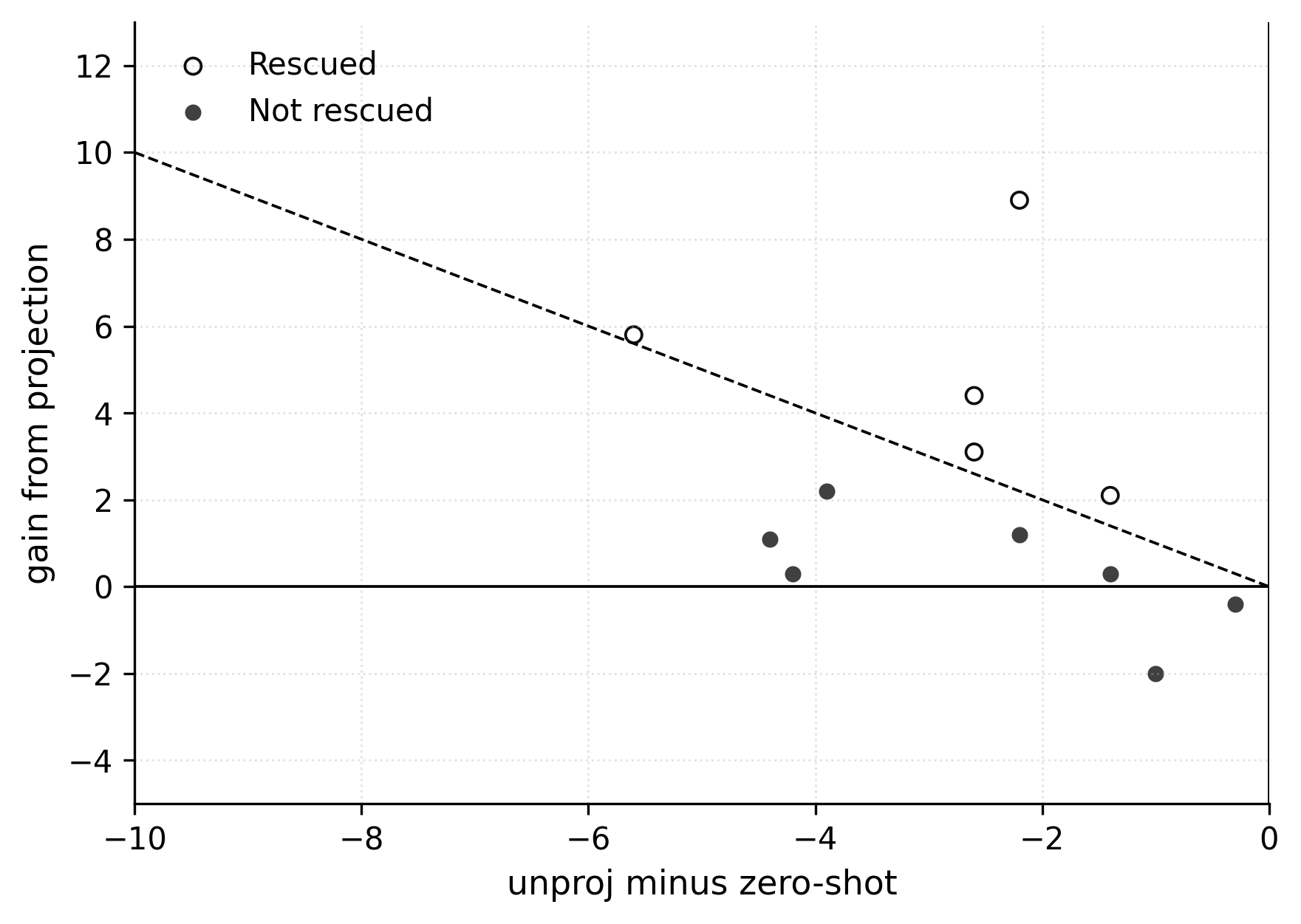}
    \caption{Projection mitigates harmful transfer}
    \label{fig: gain_projection}
\end{figure}

We conduct an ablation study to test whether intruder dimensions are truly related to degraded cross-task robustness. Concretely, for the cross-task LoRA models, we apply a projection $P_{\mathcal{S}_0}$ that constrains the low-rank update to lie in the pretrained weight subspace $\mathcal{S}_0$. The results are shown in Fig.~\ref{fig: gain_projection}. Projection often yields recovery, rescuing a nontrivial portion of harmful-transfer points, which supports our hypothesis that intruder directions contribute to reduced cross-task robustness.

Additional ablation studies on $\boldsymbol{\lambda}$-clipping and the magnitude of adaptation are provided in Appendix \ref{Ablation_Study}. A further analysis of intruder dimensions in LoRA is provided in Appendix \ref{additional_analysis_lora_intruder}, where we show that the intruder component of $\boldsymbol{\Delta}$ is often non-negligible in magnitude across layers. 

\section{Conclusion}

To mitigate forgetting during fine-tuning, this work proposes a feature-space fine-tuning framework grounded in effect–equivalence modeling, as an alternative to conventional weight-space update strategies. Through theoretical analysis, we explicitly quantify a forgetting measure and relate it to intruder dimensions. We show that VeFA introduces no intruder dimensions and therefore avoids the loss induced by them. We further derive a global bound on the total forgetting under VeFA and rigorously show that it can be explicitly controlled, demonstrating VeFA’s robustness.

Our empirical results on vision and language tasks show that VeFA can match or even outperform weight-space fine-tuning approaches while using far fewer trainable parameters. VeFA also demonstrates stronger robustness across datasets. Future research could explore alternative projected-gradient schemes that constrain the update $\boldsymbol{\Delta}$ to $\mathcal{S}_0$ while further reducing in-subspace loss. 
It could also investigate sparsity-inducing constraints on the scaling vector $\boldsymbol{\Lambda}$, connecting VeFA to sparse and mechanistic interpretability. 



\nocite{langley00}

\newpage
\bibliography{example_paper}
\bibliographystyle{icml2026}

\newpage
\appendix
\onecolumn
\section{Related Work}

\textbf{Parameter Efficient Fine Tuning (PEFT)}. PEFT aims to selectively fine-tune a small subset of parameters or incorporate lightweight trainable modules—a task that is inherently NP-hard. Existing PEFT approaches can be categorized into random approaches, rule-based approaches, and projection-based approaches based on how they choose which parameters to tune. \textbf{Randomized approaches}, such as the Random and Mixout models \cite{leemixout}, select parameters for fine-tuning without relying on task-specific data information. \textbf{Rule-based approaches} such as BitFit (\cite{zaken2021bitfit}), MagPruning (\cite{han2015learning, lee2020layer, lagunas2021block}), Adapter (\cite{houlsby2019parameter, pfeiffer2020adapterhub, ruckle2020adapterdrop}), and LoRA ( \cite{hu2022lora, karimi2021compacter, panahi2021shapeshifter} ) determine which parameters to fine-tune based on pre-defined heuristics. These methods incorporate prior knowledge to identify potentially important components in the model, thereby addressing some of the limitations of randomized approaches. However, the parameter selection process remains independent of the specific downstream data. Among rule-based PEFT methods, \textbf{LoRA} has become a de facto baseline due to its simplicity, hardware efficiency, and strong empirical performance across modalities. In our experiments, LoRA also serves as the primary baseline for comparison. \textbf{Projection-based approaches}, such as DiffPruning \cite{mallya2018piggyback, sanh2020movement, lagunas2021block} and ChildPruning (\cite{xu2021raise, mostafa2019parameter}), aim to leverage task-specific data to guide the selection of tunable parameters in a pre-trained model. Our approach also falls under the category of PEFT; however, it differs fundamentally from prior PEFT methods in that we perform fine-tuning in the feature space rather than the weight space. Unlike weight-based methods, our fine-tuning does not alter the column space of $\boldsymbol{W}_0$, thereby preserving pre-training knowledge more effectively throughout adaptation. This design is particularly advantageous when fine-tuning large-scale models (e.g., the 175-billion-parameter GPT-3), where maintaining the integrity of pre-trained representations significantly improves robustness.

\textbf{Domain Adaptation}. Although our method explicitly models and compensates for domain discrepancy, it fundamentally differs from existing domain adaptation (DA) approaches \cite{zhu2017unpaired, ganin2016domain, liang2020we, motiian2017unified}. DA methods generally rely on joint access to source and target data and seek to learn a domain-invariant representation by optimizing the feature extractor accordingly. In contrast, our problem setting assumes that source domain data is unavailable and needs to revise the pre-trained model to adapt to the downstream data. We perform feature-space transformation at the layer level, but the focus ultimately remains on parameter adaptation. Moreover, the domain discrepancy in our problem setting is substantially greater than that typically encountered in standard domain adaptation tasks. These distinctions position our work closer to fine-tuning methodologies rather than DA frameworks.

\section{Neural Parameter Geometry}
\subsection{Cosine Similarity under Neural Parameter Symmetries}
\label{cs_nps}
Neural networks exhibit rich \emph{parameter symmetries}, meaning that different parameter values can define exactly the same function \cite{navon2023equivariant,theus2025generalized}. Two examples that are particularly relevant for our analysis are permutation invariance and $\mathrm{GL}(r)$-invariance. For a two-layer MLP $f_{W_1,W_2}(x)=W_2\sigma(W_1x)$ and any permutation matrix $P$, the transformed parameters $W_1' = PW_1$ and $W_2' = W_2P^\top$ satisfy
\[
f_{W_1',W_2'}(\boldsymbol{x})
= W_2P^\top \sigma(PW_1\boldsymbol{x})
= f_{W_1,W_2}(\boldsymbol{x}),
\]
so permuting hidden units leaves the network function unchanged. Likewise, in modules that use low-rank matrix products—such as LoRA updates or the $Q,K,V,W^O$ projections in attention—a rank-$r$ factorization $UV^\top$ is invariant under the action of any invertible matrix $R\in \mathrm{GL}(r)$:
\[
UV^\top = (UR)(VR^{-\top})^\top.
\]
This $\mathrm{GL}(r)$-invariance reflects a “gauge freedom’’ in the choice of basis for the $r$-dimensional latent space: different pairs $(U,V)$ related by such transformations represent the same linear map.

As a concrete illustration of permutation invariance, consider a two-layer network (with any pointwise activation $\sigma$)
\(h = \sigma(W_1 x), \quad y = W_2 h,\) with
\[
W_1 =
\begin{pmatrix}
1 & 2\\
3 & 4
\end{pmatrix},
\quad
W_2 = \begin{pmatrix} 5 & 6 \end{pmatrix},
\quad
P =
\begin{pmatrix}
0 & 1\\
1 & 0
\end{pmatrix}.
\]
Define the permuted weights
\[
W_1' = P W_1 =
\begin{pmatrix}
3 & 4\\
1 & 2
\end{pmatrix},
\qquad
W_2' = W_2 P^\top =
\begin{pmatrix}
6 & 5
\end{pmatrix}.
\]
For any input $\boldsymbol{x} = (x_1, x_2)^\top$ we obtain
\[
h = W_1 x =
\begin{pmatrix}
x_1 + 2x_2\\
3x_1 + 4x_2
\end{pmatrix},
\qquad
h' = W_1' x =
\begin{pmatrix}
3x_1 + 4x_2\\
x_1 + 2x_2
\end{pmatrix}.
\]
The outputs of the original and permuted networks are
\[
y = W_2 h
  = 5(x_1 + 2x_2) + 6(3x_1 + 4x_2),
\qquad
y' = W_2' h'
  = 6(3x_1 + 4x_2) + 5(x_1 + 2x_2),
\]
$y' = y$ for all $\boldsymbol{x}$. Thus, the parameter pairs $(W_1,W_2)$ and $(W_1’,W_2’)$ are functionally equivalent, even though their weight matrices differ and, in particular, the maximum absolute cosine similarity between any pair of their left singular vectors is only about $0.74$. This example also shows that cosine similarity should not be used to define intruder dimensions associated with \textbf{forgetting}.

\subsection{Rank analysis of the pre-trained weight matrix}
\label{rank_analysis_pretrained}

\begin{figure}[htbp]
    \centering
    \includegraphics[width=0.8\textwidth]{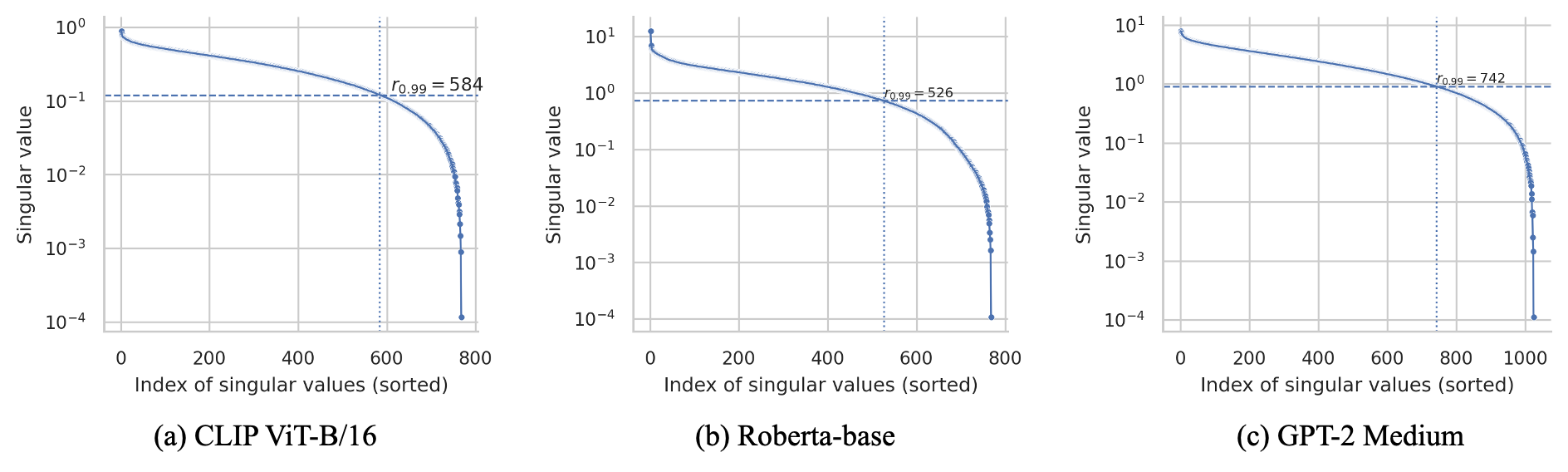}
    \caption{Spectral Decay and Effective Rank of Pre-trained Transformer Weight Matrices}
    \label{fig: rank_structure}
\end{figure}

Although in pre-trained models the weight matrices are mathematically full rank, their singular value spectra are concentrated. Most of the representational power of a pre-trained layer is carried by a low-dimensional principal subspace, while the remaining directions contribute only negligible energy. As illustrated in Fig. \ref{fig: rank_structure}, for CLIP ViT-B/16, RoBERTa-base, and GPT-2 Medium, more than 99\% of the spectral energy is already captured by roughly the top two-thirds of singular directions, with the remaining singular values quickly collapsing towards numerical noise. In our definition of intruder dimensions, we therefore restrict attention to this high-energy subspace: directions in the tiny-energy tail are poorly conditioned, highly sensitive to symmetries and optimization noise, and are unlikely to encode stable, semantically meaningful information that could be “forgotten.”

\section{Hyperparameter}
\label{appendix: hyperparameters}
\begin{table}[ht]
\centering
\footnotesize
\setlength{\tabcolsep}{6pt}
\renewcommand{\arraystretch}{1.2}
\caption{Hyper-parameter settings for RoBERTa-Base and RoBERTa-Large on GLUE benchmark tasks.}
\label{hyperparameter_glue}
\begin{tabular}{c|c|c|c|c|c|c|c}
\toprule
\textbf{Model} & \textbf{Hyper-Parameters} & \textbf{SST-2} & \textbf{MRPC} & \textbf{CoLA} & \textbf{QNLI} & \textbf{RTE} & \textbf{STS-B} \\
\midrule
 & Optimizer          & \multicolumn{6}{c}{AdamW} \\
 & Warmup Ratio       & \multicolumn{6}{c}{0.06} \\
 & LR Schedule        & \multicolumn{6}{c}{Linear} \\
 \midrule
 \multirow{7}{*}{\rotatebox{90}{\textbf{Base}}}
 & Epochs             & 60  & 30  & 80  & 25  & 80  & 50 \\
 & Learning Rate (VeFA) & 1E-2 & 4E-3 & 1E-2 & 1E-2 & 4E-3 & 4E-3 \\
 & Weight Decay (VeFA)  & 0.05 & 0.01   & 0   & 0.01 & 0.1 & 0.1 \\
 & Learning Rate (Head) & 4E-3 & 4E-3 & 1E-2 & 4E-3 & 4E-3 & 4E-3 \\
 & Weight Decay (Head)  & 0.05 & 0.01 & 0.01 & 0.05 & 0.1 & 0.1 \\
 & Max Seq. Len.        & \multicolumn{6}{c}{512} \\
 & Batch Size           & \multicolumn{6}{c}{64} \\
\midrule
\multirow{8}{*}{\rotatebox{90}{\textbf{Large}}}
 & Epochs             & 40  & 20  & 20  & 25  & 20  & 20 \\
 & Learning Rate (VeFA) & 1E-2 & 2E-2 & 1E-2 & 1E-2 & 2E-2 & 4E-3 \\
 & Weight Decay (VeFA)  & 0.1  & 0.1  & 0    & 0.01 & 0.1  & 0.1 \\
 & Learning Rate (Head) & 6E-3 & 4E-3 & 4E-3 & 4E-4 & 4E-3 & 4E-3 \\
 & Weight Decay (Head)  & 0.1  & 0.1  & 0.01 & 0.01 & 0.1  & 0.1 \\
 & Max Seq. Len.        & \multicolumn{6}{c}{512} \\
 & Batch Size           & \multicolumn{6}{c}{32} \\
\bottomrule
\end{tabular}
\end{table}

\begin{table}[H]
\centering
\caption{Hyperparameter configurations for VeFA on the E2E benchmark, for GPT-2 Medium and Large models.}
\begin{tabular}{lcc}
\toprule
\textbf{Hyperparameter} & \textbf{Medium} & \textbf{Large} \\
\midrule
\# GPUs            & 1        & 1        \\
Optimizer          & \multicolumn{2}{c}{AdamW} \\
Learning Rate Schedule & \multicolumn{2}{c}{Linear} \\
Weight Decay       & \multicolumn{2}{c}{0.01} \\
Batch Size         & \multicolumn{2}{c}{8} \\
Epochs             & \multicolumn{2}{c}{5} \\
Warmup Steps       & \multicolumn{2}{c}{500} \\
Label Smooth       & \multicolumn{2}{c}{0.1} \\
Learning Rate      & 3E-2     & 2E-2     \\
\bottomrule
\end{tabular}
\end{table}

\section{Proof of Theoretical Analysis}
\subsection{Proof of Proposition \ref{intruder_dimension_loss}}
\label{proof_intruder_dimension_loss}
    Since $P_{\mathcal{S}_0}^\perp$ is the projection operator admitting $P_{\mathcal{S}_0}^\perp = P_{\mathcal{S}_0}^\perp P_{\mathcal{S}_0}^\perp$, the above inequality can be further written as:
    $$
    \begin{aligned}
       \text{Tr} \left( P_{\mathcal{S}_0}^\perp \Delta \Sigma_{\boldsymbol{x}} \Delta^T {P_{\mathcal{S}_0}^\perp}^T \right)
         &= \sum_i \sigma_i \left( \Sigma_{\boldsymbol{x}} \right) \cdot \Big\Vert \left[ P_{\mathcal{S}_0}^\perp \Delta Q \right]_{\cdot, i} \Big\Vert_2^2 \\
        & \geq \sigma_{\min} \left( \Sigma_{\boldsymbol{x}} \right) \sum_i \Big\Vert \left[ P_{\mathcal{S}_0}^\perp \Delta Q \right]_{\cdot, i} \Big\Vert_2^2 \\
        & = \sigma_{\min} \left( \Sigma_{\boldsymbol{x}} \right) \sum_i \Big\Vert P_{\mathcal{S}_0}^\perp \Delta Q \Big\Vert_F^2 \\ 
        & = \sigma_{\min} \left( \Sigma_{\boldsymbol{x}} \right) \big\Vert P_{\mathcal{S}_0}^\perp \Delta \big\Vert_F^2,
    \end{aligned}
    $$
    in which $Q$ is a unitary/conjugate matrix in the SVD of $\Sigma_{\boldsymbol{x}}$ (or equivalently orthogonal diagonalization for such symmetric matrix), and thus $Q$ is orthogonal and is canceled in the Frobenius norm. 

    If we write the weight change $\Delta$ as the difference between $\boldsymbol{W}_{\mathrm{ft}}$ and $\boldsymbol{W}_0$, the F-norm component admits:
    $$
    \big\Vert P_{\mathcal{S}_0}^\perp \Delta \big\Vert_F \geq \big\Vert P_{\mathcal{S}_0}^\perp \boldsymbol{W}_{\mathrm{ft}} \big\Vert_F - \big\Vert P_{\mathcal{S}_0}^\perp \boldsymbol{W}_0 \big\Vert_F.
    $$
    in which:
    $$
    \begin{aligned}
        P_{\mathcal{S}_0}^\perp \boldsymbol{W}_{\mathrm{ft}} & = P_{\mathcal{S}_0}^\perp \sum_{i = 1}^r \sigma_{{\mathrm{ft}},i} \boldsymbol{u}_{{\mathrm{ft}},i} \boldsymbol{v}_{{\mathrm{ft}},i}^T \\
        & = \sum_{i = 1}^r \sigma_{{\mathrm{ft}},i} \left( P_{\mathcal{S}_0}^\perp \boldsymbol{u}_{{\mathrm{ft}},i} \right) \boldsymbol{v}_{{\mathrm{ft}},i}^T 
    \end{aligned}
    $$
    and further:
    $$
    \begin{aligned}
        \big\Vert P_{\mathcal{S}_0}^\perp \boldsymbol{W}_{\mathrm{ft}} \big\Vert_F^2
        & = \text{Tr} \left( P_{\mathcal{S}_0}^\perp \boldsymbol{W}_{\mathrm{ft}} (P_{\mathcal{S}_0}^\perp \boldsymbol{W}_{\mathrm{ft}})^T \right) \\
        & = \sum_{i = 1}^r \sigma_{{\mathrm{ft}},i}^2 \big\Vert P_{\mathcal{S}_0}^\perp \boldsymbol{u}_{{\mathrm{ft}},i} \big\Vert_2^2
    \end{aligned}
    $$
    From the definition of intruder dimension above, it is straightforward that the projection of vector $\boldsymbol{u}_{{\mathrm{ft}},i}$ onto the complement of $\mathcal{S}_0$ will have an upper bound $1-\varepsilon$, and therefore:
    $$
    \big\Vert P_{\mathcal{S}_0}^\perp \boldsymbol{W}_{\mathrm{ft}} \big\Vert_F^2 > (1 - \varepsilon) \sum_{i = 1}^r \sigma_{{\mathrm{ft}},i}^2 \geq (1 - \varepsilon) r_{\mathrm{ft}} \sigma_{{\mathrm{ft}}, r_{\mathrm{ft}}}^2.
    $$
    On the other hand, $\big\Vert P_{\mathcal{S}_0}^\perp \boldsymbol{W}_0 \big\Vert_F^2 = \sum_{i = r_0 + 1}^r \sigma_{0,i}^2 \big\Vert P_{\mathcal{S}_0}^\perp \boldsymbol{u}_{0,i} \big\Vert_2^2 = \sum_{i = r_0 + 1}^r \sigma_{0,i}^2$ measures the residual portion of spectral power:
    $$
    \big\Vert P_{\mathcal{S}_0}^\perp \boldsymbol{W}_0 \big\Vert_F^2 \leq (1 - \alpha) \Vert \boldsymbol{W}_0 \Vert_F^2.
    $$
    Combining the two parts together, we can then derive the bound:
    $$
    \begin{aligned}
    \Delta \mathcal{L} & \geq \sigma_{\min} \left( \Sigma_{\boldsymbol{x}} \right) \big\Vert P_{\mathcal{S}_0}^\perp \Delta \big\Vert_F^2 \\
    & \geq \sigma_{\min} \left( \Sigma_{\boldsymbol{x}} \right) \left[ \sqrt{(1 - \varepsilon) r_{\mathrm{ft}} \sigma_{{\mathrm{ft}}, r_{\mathrm{ft}}}^2} - \sqrt{1 - \alpha} \Vert \boldsymbol{W}_0 \Vert_F \right]_+^2
    \end{aligned}
    $$

\subsection{Proof of Proposition \ref{in_subspace_loss}}
\label{proof_in_subspace_loss}
Recall that under VeFA, the fine-tuned weight matrix is
\(
\boldsymbol{W}_{\mathrm{ft}} = \boldsymbol{W}_0 + \boldsymbol{W}_0\boldsymbol{\Lambda}_k
\),
where \(\boldsymbol{\Lambda}_k = \operatorname{diag}(\boldsymbol{k})\).
Hence the update is
\(\boldsymbol{\Delta}
= \boldsymbol{W}_0 \boldsymbol{\Lambda}_k\).

Since \(\operatorname{Col}(\boldsymbol{\Delta}) \subseteq \operatorname{Col}(\boldsymbol{W}_0)\subseteq
\mathcal{S}_0\), we have \(P_{\mathcal{S}_0}\boldsymbol{\Delta} = \boldsymbol{\Delta}\) and therefore
\(\operatorname{Tr}\!\bigl(P_{\mathcal{S}_0}\boldsymbol{\Delta}
\boldsymbol{\Sigma}_{\boldsymbol{x}}\boldsymbol{\Delta}^\top
P_{\mathcal{S}_0}^\top\bigr)
=
\operatorname{Tr}\!\bigl(\boldsymbol{\Delta}
\boldsymbol{\Sigma}_{\boldsymbol{x}}\boldsymbol{\Delta}^\top\bigr)\).

We now bound this term. Because
\(\boldsymbol{\Sigma}_{\boldsymbol{x}}\) and
\(\boldsymbol{\Delta}\boldsymbol{\Delta}^\top\) are PSD,
\[
\operatorname{Tr}\!\bigl(\boldsymbol{\Delta}
\boldsymbol{\Sigma}_{\boldsymbol{x}}\boldsymbol{\Delta}^\top\bigr)
=
\operatorname{Tr}\!\bigl(\boldsymbol{\Sigma}_{\boldsymbol{x}}
\boldsymbol{\Delta}^\top\boldsymbol{\Delta}\bigr)
\le
\|\boldsymbol{\Sigma}_{\boldsymbol{x}}\|_2
\operatorname{Tr}\!\bigl(\boldsymbol{\Delta}^\top\boldsymbol{\Delta}\bigr)
=
\|\boldsymbol{\Sigma}_{\boldsymbol{x}}\|_2
\|\boldsymbol{\Delta}\|_F^2 .
\]

Next, using sub-multiplicativity of matrix norms, we obtain
\[
\|\boldsymbol{\Delta}\|_F
=
\|\boldsymbol{W}_0 \boldsymbol{\Lambda}_k\|_F
\le
\|\boldsymbol{W}_0\|_F \,\|\boldsymbol{\Lambda}_k\|_2
=
\|\boldsymbol{W}_0\|_F \,\max_j |k_j|.
\]
Thus
\[
\|\boldsymbol{\Delta}\|_F^2
\le
\|\boldsymbol{W}_0\|_F^2
\bigl(\max_j |k_j|\bigr)^2 .
\]

Combining the two bounds yields
\[
\operatorname{Tr}\!\bigl(P_{\mathcal{S}_0}\boldsymbol{\Delta}
\boldsymbol{\Sigma}_{\boldsymbol{x}}\boldsymbol{\Delta}^\top
P_{\mathcal{S}_0}^\top\bigr)
=
\operatorname{Tr}\!\bigl(\boldsymbol{\Delta}
\boldsymbol{\Sigma}_{\boldsymbol{x}}\boldsymbol{\Delta}^\top\bigr)
\le
\|\boldsymbol{\Sigma}_{\boldsymbol{x}}\|_2
\|\boldsymbol{W}_0\|_F^2
\bigl(\max_j |k_j|\bigr)^2,
\]
which is exactly the claimed inequality in Proposition \ref{in_subspace_loss}.

\section{Additional Ablation Study}
\label{Ablation_Study}

\textbf{Ablation on $\lambda$-Clipping}. We first examine the effect of $\lambda$-clipping on VeFA’s performance. On the E2E benchmark, we consider two ablation settings: VeFA with explicit clipping and VeFA without clipping. The results are reported in Tab.~\ref{Tab: E2E_ablation}. Overall, the two settings yield comparable performance, indicating that clipping does not noticeably reduce expressive power, even on more complex tasks.

\begin{table}[t]
\centering
\caption{Performance Comparison of VeFA With and Without Clipping on the E2E Benchmark}
\label{Tab: E2E_ablation}
\setlength{\tabcolsep}{2pt}
\renewcommand{\arraystretch}{1.3}

\begin{tabular}{c|c|c|c|c|c|c|c}
\toprule
\centering
 & \textbf{Type} & \textbf{\# Params} & \textbf{BLEU} & \textbf{NIST} &
   \textbf{MET} & \textbf{ROUGE-L} & \textbf{CIDEr} \\
\midrule
\multirow{2}{*}{\rotatebox{90}{\textbf{Med}}}
 & VeFA (w/o clipping) & 0.073M & 66.42 & 8.5852 & 44.40 & 66.46 & 2.2134 \\
 & VeFA (w/ clipping) & 0.073M & 66.35 & 8.5628 & 44.79 & 66.93 & 2.2274 \\
\midrule
\multirow{2}{*}{\rotatebox{90}{\textbf{Large}}}
 & VeFA (w/o clipping) & 0.14M & 67.62 & 8.5956 & 46.04 & 68.77 & 2.3858 \\
 & VeFA (w/ clipping) & 0.14M & 67.38 & 8.5889 & 46.11 & 68.86 & 2.3684 \\
\bottomrule
\end{tabular}
\end{table}

\begin{figure}[t]
    \centering
    \begin{subfigure}[t]{0.45\linewidth}
        \centering
        \includegraphics[width=\linewidth]{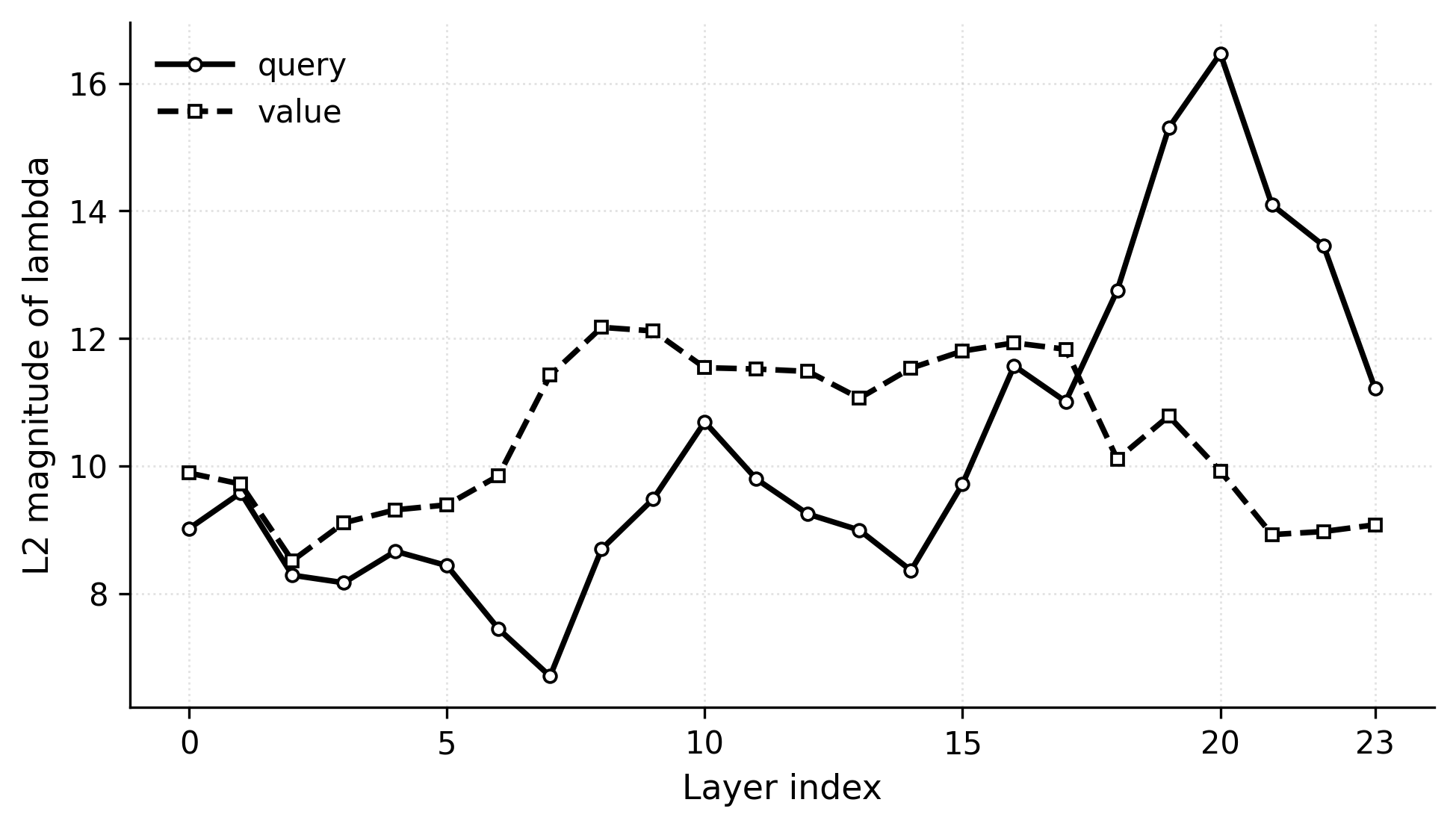}
        \caption{RTE}
    \end{subfigure}
    \hfill
    \begin{subfigure}[t]{0.45\linewidth}
        \centering
        \includegraphics[width=\linewidth]{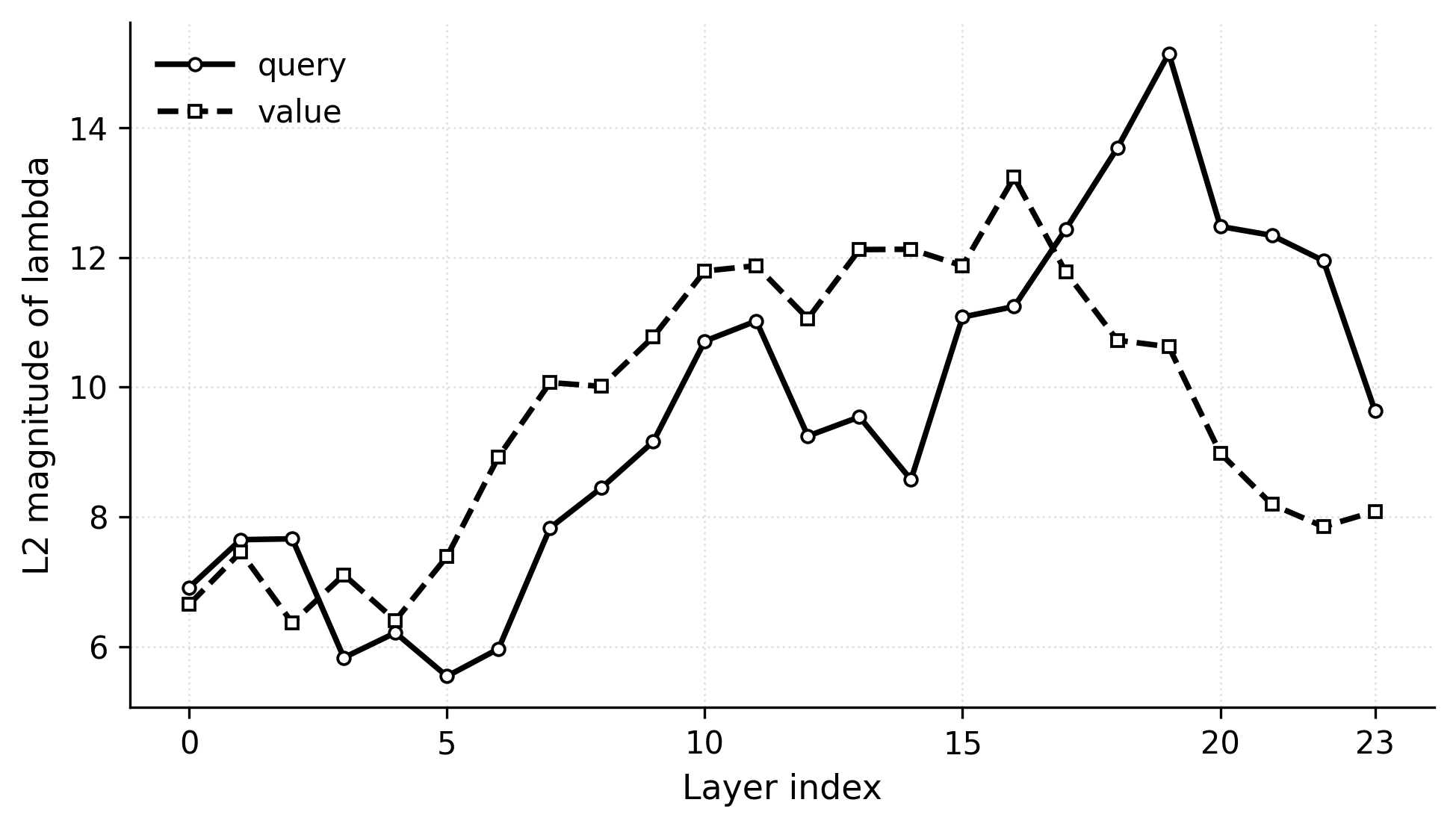}
        \caption{MRPC}
    \end{subfigure}
    \caption{Magnitude of the adapted $\boldsymbol{\lambda}$ vector for query and value matrices across layers for RoBERTa-L.}
    \label{fig: magnitude_adaptation}
\end{figure}

\textbf{Magnitude of Adaptation}. In Fig. \ref{fig: magnitude_adaptation}, we visualize the layer-wise magnitude of the learned $\boldsymbol{\lambda}$ vectors after fine-tuning on RTE and MRPC. Overall, the largest adaptation happens for query matrices compared to the value ones, indicating a larger need or ease for finetuning a model there. Moreover, consistent with observations in prior parameter-efficient adaptation studies \cite{kopiczko2023vera,zhang2023adalora,liu2022p}, we observe stronger adaptations in later layers than in earlier layers.

\section{Additional Numerical Experiments}

\subsection{Comparison of Weight-Space and Feature-Space Adaptation}
\label{numerical_example_weight_feature}
\begin{figure}[htbp]
    \centering
    \includegraphics[width=0.8\textwidth]{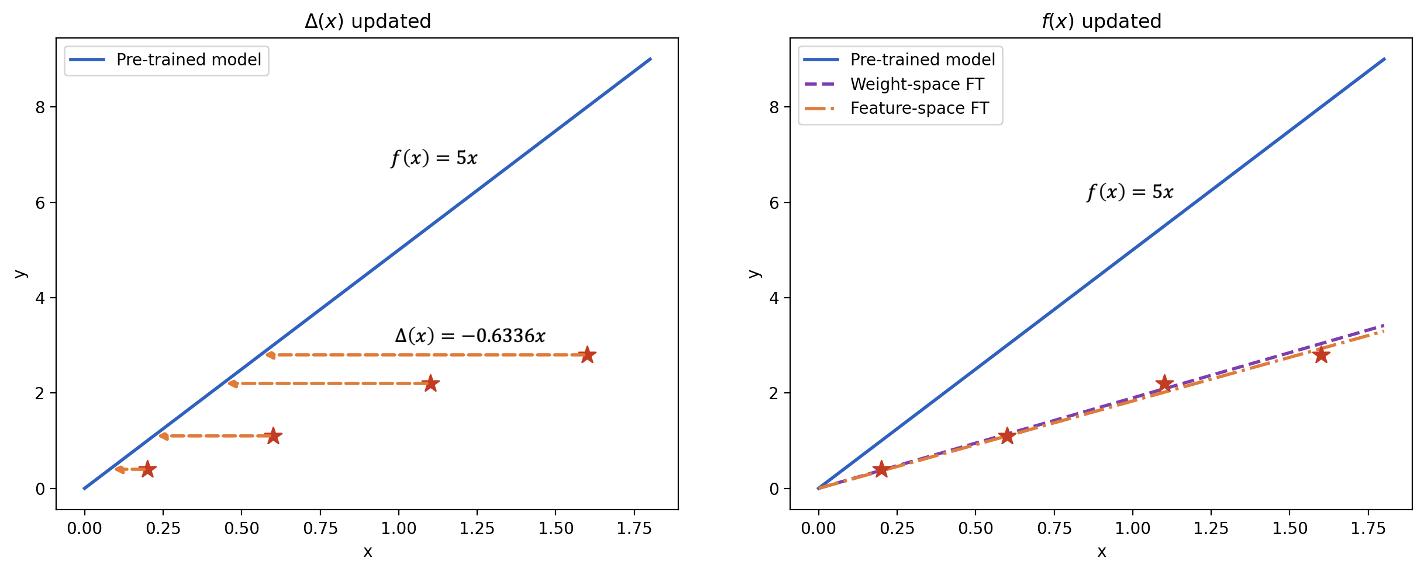}
    \caption{Comparison of weight-space fine-tuning  and feature-space fine-tuning}
    \label{fig: comparison_1d}
\end{figure}

To better understand the differences between feature, we then give a simple one-dimensional example. Suppose that the pre-trained model is $y = 5x$, and the downstream dataset consists of 
$\{[0.2, 0.4], [0.6, 1.1], [1.1, 2.2], [1.6, 2.8]\}$. 
Using gradient descent, we estimate $\delta_1$ in  $y = (5 + \delta_1)x$ (weight-space adaptation) and $\delta_2$ in $y = 5(x + \delta_2x)$ (feature-space adaptation). The final results are shown in the Fig. \ref{fig: comparison_1d}. In weight-space fine-tuning, the model adapts by directly updating the parameters, which shifts the regression line (purple dashed) away from the pre-trained function $f(x) = 5x$ in second sub-figure. In feature-space fine-tuning, the pre-trained model is kept frozen and adaptation is achieved by learning a lightweight transformation $\Delta(x)$ applied to the input features (orange dashed line in first sub-figure), resulting in the orange dashed regression line in second sub-figure. 

\subsection{Effectiveness on the Visual Classification Benchmark}
\label{Complete effectivess_visual}
Few-shot learning closely matches the setting studied in this paper, where {downstream data are much smaller in scale than the pre-training datasets and the goal is to mitigate overfitting and forgetting during fine-tuning}. We compare VeFA with several prompt-based methods, including CoOp (4) with four learnable context tokens \cite{zhou2022learning}, CoCoOp \cite{zhou2022conditional}, PLOT++ \cite{chen2022plot}, and ProGrad \cite{zhu2023prompt} with 16 learnable tokens, which is conceptually related to our approach in that it constrains updates in the few-shot regime. We also report results for adapter-based methods, including TIP-Adapter-F \cite{zhang2022tip}, where we reduce the validation set to a reasonable size, TaskRes \cite{yu2023task} and CLIP-Adapter.

The final comparison is reported in Tab.\ref{Tab: visual_complete}. Our VeFA outperforms all eight baselines across seven datasets, validating the effectiveness of VeFA.

\begin{table*}[t]
\caption{Performance comparison between VeFA and eight baselines across different datasets under zero-shot, 1-shot, 4-shot, 16-shot fine-tuning. The visual backbone is ViT-B/16. Best results are in \textbf{bold}.}
\centering
\small
\setlength{\tabcolsep}{4.2pt}
\begin{tabular}{c|c| c|c|c|c|c|c|c|c}
\toprule
Shots & \textbf{Type} & \textbf{Caltech101} & \textbf{Food101} & \textbf{Oxford Pets} & \textbf{Oxford Flowers} & \textbf{EuroSAT} & \textbf{DTD} & \textbf{UCF101} & \textbf{Avg} \\
\midrule
\multirow{1}{*}{0}
& CLIP & 93.2 & 85.2 & 88.2 & 67.3 & 42.3 & 44.3 & 65.0 & 69.4 \\
\midrule

\multirow{9}{*}{1}
& CoOp (4)
& 93.2 & 82.6 & 90.3 & 72.7 & 50.9 & 50.1 & 70.7 & 72.9 \\
& CoCoOp 
& 94.1 & 84.9 & 91.9 & 73.4 & 55.4 & 52.6 & 70.4 & 74.7 \\
& TIP-Adapter-F
& 94.0 & 85.8 & 90.6 & 83.8 & 67.8 & 51.6 & 73.4 & 78.1 \\
& CLIP-Adapter
& 92.0 & 86.1 & 89.0 & 71.3 & 49.3 & 44.2 & 66.9 & 71.3 \\
& PLOT++ 
& \textbf{94.3} & 86.2 & 91.9 & 80.5 & 65.4 & 54.6 & 74.3 & 78.2 \\
& TaskRes
& 93.6 & 84.6 & 90.2 & 81.7 & 65.4 & 53.8 & 71.7 & 77.3 \\
& ProGrad
& 93.5 & 84.9 & 91.4 & 80.9 & 57.0 & 52.8 & 73.3 & 76.3 \\
& LoRA & 94.0 & 82.3 & 91.8 & 79.4 & 71.8 & 53.0 & \textbf{75.4} & 78.2\\
& VeFA (Ours) & 94.1 & \textbf{86.3} & \textbf{93.3} & \textbf{84.3} & \textbf{73.3} & \textbf{55.0} & 74.2 & \textbf{80.1} \\
\midrule

\multirow{9}{*}{4}
& CoOp (4)
& 94.5 & 83.5 & 92.3 & 86.6 & 65.8 & 58.5 & 78.1 & 79.9 \\
& CoCoOp 
& 94.8 & 86.3 & 92.7 & 81.5 & 61.7 & 55.7 & 75.3 & 78.3 \\
& TIP-Adapter-F
& 94.8 & \textbf{86.5} & 91.9 & 92.1 & 76.8 & 59.8 & 78.1 & 82.9\\
& CLIP-Adapter
& 94.0 & 86.5 & 90.8 & 73.1 & 51.2 & 46.1 & 70.6 & 73.2 \\
& PLOT++ 
& 95.1 & 86.5 & 92.6 & 92.9 & 83.2 & 62.4 & 79.8 & 84.6 \\
& TaskRes 
& 95.0 & 86.0 & 91.9 & 85.0 & 74.2 & 60.1 & 76.2 & 81.2 \\
& ProGrad 
& 94.4 & 85.4 & 92.1 & 91.1 & 69.6 & 59.7 & 77.9 & 81.5 \\
& LoRA & 95.2 & 81.6 & 90.8 & \textbf{93.8} & \textbf{87.5} & 62.2 & 78.5 & 84.2\\
& VeFA (Ours) & \textbf{95.6} & 86.0 & \textbf{93.4} & 93.0 & 87.1 & \textbf{65.7} & \textbf{80.2} & \textbf{85.7} \\
\midrule

\multirow{9}{*}{16}
& CoOp (4)
& 95.5 & 85.1 & 92.4 & 96.4 & 83.5 & 69.2 & 81.9 & 86.3 \\
& CoCoOp 
& 95.1 & 87.4 & 93.4 & 89.1 & 73.6 & 63.7 & 77.2 & 82.3 \\
& TIP-Adapter-F 
& 95.7 & 86.8 & 92.6 & 96.2 & 85.9 & 70.8 & 83.9 & 87.4 \\
& CLIP-Adapter
& 94.9 & 87.1 & 92.3 & 92.9 & 71.4 & 59.4 & 80.2 & 82.6 \\
& PLOT++ 
& 96.0 & 87.1 & 93.6 & 97.6 & \textbf{92.0} & 71.4 & 85.3 & 89.0 \\
& TaskRes 
& 95.8 & 86.9 & 92.4 & 97.5 & 82.7 & 71.5 & 84.0 & 87.3 \\
& ProGrad
& 95.9 & 85.8 & 92.8 & 96.6 & 83.6 & 68.8 & 82.7 & 86.6 \\
& LoRA  & 96.3 & 84.0 & 91.3 & \textbf{98.2} & 90.2 & 71.8 & 86.2& 88.3\\
& VeFA (Ours)  & \textbf{96.4} & \textbf{87.8} & \textbf{94.4} & 97.5 & 91.3 & \textbf{72.5} & \textbf{86.4} & \textbf{89.5}\\
\bottomrule
\end{tabular}
\label{Tab: visual_complete}
\end{table*}

\subsection{Robustness on the Visual Classification Benchmark}
\label{Complete robustness_visual}

\begin{figure}[t]
    \centering
    \begin{subfigure}[t]{0.48\linewidth}
        \centering
        \includegraphics[width=\linewidth]{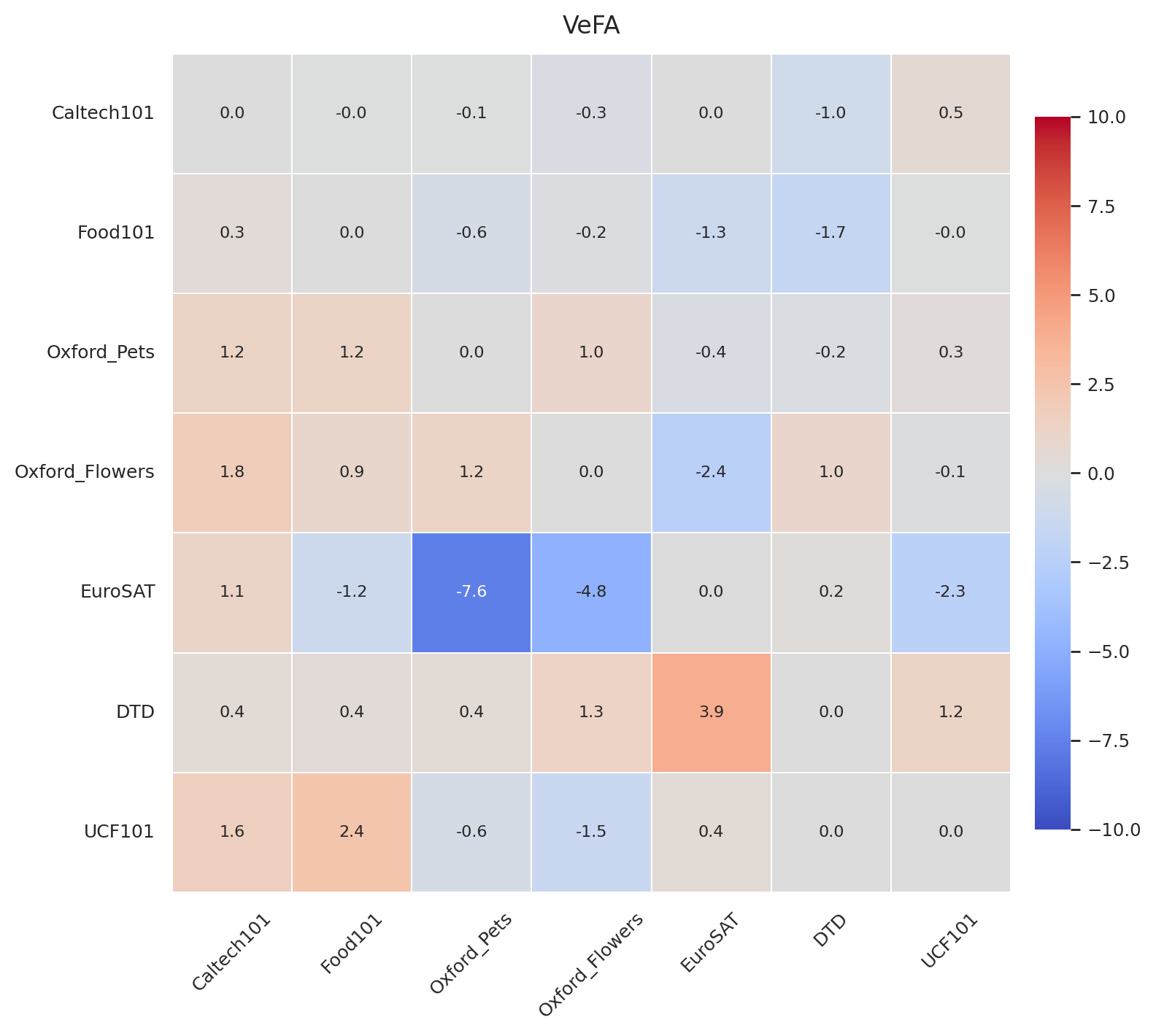}
        \caption{VeFA}
        \label{fig:vefa_heatmap}
    \end{subfigure}
    \hfill
    \begin{subfigure}[t]{0.48\linewidth}
        \centering
        \includegraphics[width=\linewidth]{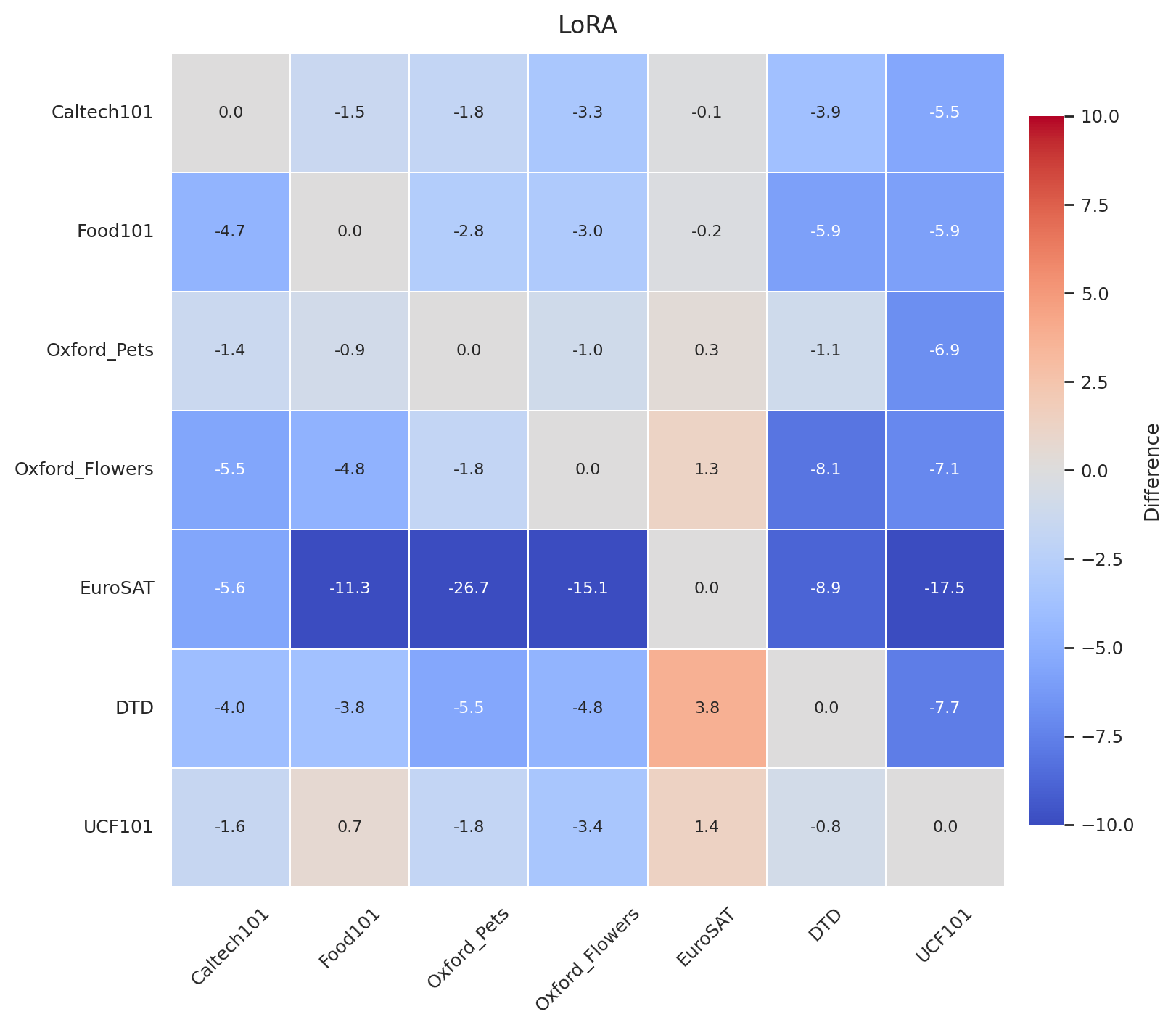}
        \caption{LoRA}
        \label{fig:lora_heatmap}
    \end{subfigure}
    \caption{Cross-dataset transfer heatmaps. Each entry shows the performance change relative to zero-shot learning (diagonal entries are set to 0).}
    \label{fig:heatmap_vefa_lora}
\end{figure}

Figure \ref{fig:heatmap_vefa_lora} shows a clear difference in cross-dataset transfer behavior between VeFA and LoRA. VeFA exhibits much more stable transfer, with most entries staying close to zero and several directions even showing small positive gains, indicating that adaptation causes limited disruption to the pre-trained representations. In contrast, LoRA is dominated by negative shifts, with many strongly negative cells and several severe drops (e.g., when adapting on EuroSAT), suggesting a higher risk of harmful transfer and forgetting. Overall, VeFA consistently reduces both the frequency and magnitude of performance degradation across datasets, demonstrating more reliable and robust cross-domain generalization.

\subsection{Effectiveness on the GLUE Benchmark}
\label{Complete effectivess_GLUE}

A broad comparison of VeFA against full fine-tuning, BitFit, adapter-based methods, LoRA, and its variants is shown in Tab.\ref{Tab: GLUE_complete}. VeFA achieves comparable performance—and often outperforms many alternatives—while using the fewest trainable parameters.
\begin{table*}[t]
\caption{Performance on GLUE with different fine-tuning methods. We report Matthews correlation
for CoLA, Pearson correlation for STS-B, and accuracy for all other tasks; higher is better.
Results for all methods except VeFA and LoRA are taken from prior work
\cite{hu2022lora,kopiczko2023vera, zhang2023lora}.}
\label{Tab: GLUE_complete}
\centering

\setlength{\tabcolsep}{5pt}
\renewcommand{\arraystretch}{1.1}

\begin{footnotesize}

\begin{tabular}{c|c|c|c|c|c|c|c|c|c}
\toprule
 & \textbf{Type} & \textbf{\# Params} & \textbf{SST-2} & \textbf{MRPC} & \textbf{CoLA} &
   \textbf{QNLI} & \textbf{RTE} & \textbf{STS-B} & \textbf{AVG} \\
\midrule
\multirow{8}{*}{\rotatebox{90}{\textbf{Base}}}
 & FT     & 125M    & 94.8 & 90.2 & 63.6      & 92.8 & 78.7 & 91.2 & 85.2 \\
 & BitFit & 0.1M    & 93.7 & 92.7 & 62.0      & 91.8 & 81.5 & 90.8 & 85.4 \\
 & $\text{Adpt}^{\text{D}}$   & 0.3M    & 94.2$\pm$0.1 & 88.5$\pm$1.1 & 60.8$\pm$0.4 & 93.1$\pm$0.1 & 71.5$\pm$2.7 & 89.7$\pm$0.3 & 83.0 \\
  & $\text{Adpt}^{\text{D}}$   & 0.9M    & 94.7$\pm$0.3 & 88.4$\pm$0.1 & 62.6$\pm$0.9 & 93.0$\pm$0.2 & 75.9$\pm$2.2 & 89.3$\pm$0.1 & 84.2 \\
 & LoRA   & 0.3M    & 94.7$\pm$0.2 & 90.1$\pm$0.9 & 64.7$\pm$1.6 & 92.6$\pm$0.2 &  86.6$\pm$0.9 & 91.4$\pm$0.1 & \textbf{86.7} \\
  & LoRA-FA & 1.8M & 94.8& 90.0& 63.6& 92.5& 67.9& 89.6& 83.1 \\
 & VeRA   & 0.043M  & 94.6$\pm$0.1 & 89.5$\pm$0.5 & 65.6$\pm$0.8 & 91.8$\pm$0.2 &
              78.7$\pm$0.7 & 90.7$\pm$0.2 & 85.2 \\
 & VeFA (Ours)   & \textbf{0.018M} & 94.1$\pm$0.1 & 88.5$\pm$0.7 & 63.3$\pm$1.0 &
              91.7$\pm$0.2 & 83.0$\pm$0.4 & 90.7$\pm$0.1 & 85.2 \\
\midrule
\multirow{9}{*}{\rotatebox{90}{\textbf{Large}}}
 & FT & 355M & 96.4& 90.9& 68.0& 94.7& 86.6& 92.4& 88.1\\
& $\text{Adpt}^{\text{P}}$   & 3M    & 96.1$\pm$0.3 & 90.2$\pm$0.7 & 68.3$\pm$1.0 & 94.8$\pm$0.2 & 83.8$\pm$2.9 & 92.1$\pm$0.7 & 87.6 \\
& $\text{Adpt}^{\text{P}}$   & 0.8M    & 96.6$\pm$0.2 & 89.7$\pm$1.2 & 67.8$\pm$2.5 & 94.8$\pm$0.3 & 80.1$\pm$2.9 & 91.9$\pm$0.4 & 86.8 \\
& $\text{Adpt}^{\text{H}}$   & 6M    & 96.2$\pm$0.3 & 88.7$\pm$2.9 & 66.5$\pm$4.4 & 94.7$\pm$0.2 & 83.4$\pm$1.1 & 91.0$\pm$1.7 & 86.8 \\
& $\text{Adpt}^{\text{H}}$   & 0.8M    & 96.3$\pm$0.5 & 87.7$\pm$1.7 & 66.3$\pm$2.0 & 94.7$\pm$0.2 & 72.9$\pm$2.9 & 91.5$\pm$0.5 & 84.9 \\
 & LoRA   & 0.8M    & 96.2$\pm$0.2 & 89.7$\pm$1.1 & 66.0$\pm$2.0 &
              94.8$\pm$0.3 & 90.2$\pm$0.6 & 92.4$\pm$0.3 & \textbf{88.2} \\
& LoRA-FA & 3.7M& 96.0& 90.0 & 68.0& 94.4& 86.1& 92.0& 87.7\\
 & VeRA   & 0.061M  & 96.1$\pm$0.1 & 90.9$\pm$0.7 & 68.0$\pm$0.8 &
              94.4$\pm$0.2 & 85.9$\pm$0.7 & 91.7$\pm$0.8 & 87.8 \\
 & VeFA (Ours)  & \textbf{0.049M} & 95.8$\pm$0.2 & 90.4$\pm$0.8 & 68.0$\pm$1.2 &
              94.1$\pm$0.3 & 87.0$\pm$0.4 & 91.2$\pm$0.4 & 87.8 \\
\bottomrule
\end{tabular}
\end{footnotesize}
\end{table*}

\subsection{Robustness on the GLUE Benchmark}
\label{Complete robustness_GLUE}

Table \ref{Tab: GLUE_robustness} summarizes cross-task robustness on GLUE under both RoBERTa-base (B) and RoBERTa-large (L).
Overall, \emph{VeFA consistently achieves markedly higher robustness than LoRA across all robustness settings}, while using \emph{around $16\times$ fewer trainable parameters} (0.018M vs.\ 0.3M).
On RoBERTa-base, VeFA improves the average robustness from 4.0 to 18.8 for $\tau=1$ and from 4.3 to 17.2 for $\tau=10$. It also outperforms LoRA under $\bar{R}$ (6.5 to 11.3) and FedAvg (5.48 to 8.53). The gains are particularly pronounced on STS-B (e.g., 44.9--46.9 for VeFA vs.\ 6.1--9.5 for LoRA), indicating substantially better transferability across tasks.
The trend is even clearer on RoBERTa-large, where LoRA's average robustness is close to zero or even negative in several settings (e.g., 1.5 and 1.3 for $\tau\in\{1,10\}$), while VeFA maintains strong positive robustness (about 20.0 and 19.6). VeFA also remains consistently strong under the other merging strategies on RoBERTa-large (e.g., 18.47 with FedAvg and 18.7 with $\bar{R}$), suggesting that feature-space adaptation yields adapters that are more compatible for cross-task reuse and combination than weight-space LoRA updates.

\begin{table}[t]
\caption{Robustness on GLUE benchmark. We report Matthews correlation for CoLA, Pearson correlation for STS-B, and accuracy for all other tasks; higher is better.}
\label{Tab: GLUE_robustness}
\centering

\setlength{\tabcolsep}{5pt}
\renewcommand{\arraystretch}{1.1}

\begin{footnotesize}

\begin{tabular}{c|c|c|c|c|c|c|c|c|c}
\toprule
& \textbf{Type} & \textbf{\# Params} & \textbf{SST-2} & \textbf{MRPC} & \textbf{CoLA} &
   \textbf{QNLI} & \textbf{RTE} & \textbf{STS-B} & \textbf{AVG} \\
   \midrule 
\multirow{8}{*}{\rotatebox{90}{\textbf{Base}}} & LoRA ($\tau=1$) & 0.3 M & 15.4$\pm$0.9 & 0& 0.1$\pm$4.8& 2.3$\pm$1.5& 0& 6.1$\pm$7.1& 4.0\\
& VeFA ($\tau=1$) & 0.018 M & 4.8$\pm$1.4 & 10.1$\pm$8.6 & 37.5$\pm$8.6  & 2.8$\pm$1.3 & 12.6$\pm$4.5& 44.9$\pm$3.9& \textbf{18.8}\\
& LoRA ($\tau=10$)& 0.3 M & 16.0$\pm$1.1 & 0 & -2.7$\pm$1.1 & 3.0$\pm$0.5 & 0 & 9.5$\pm$2.5& 4.3\\
& VeFA ($\tau=10$)& 0.018 M & 11.4$\pm$8.7 & 8.7$\pm$2.5 & 21.0$\pm$23.0 & 2.3$\pm$0.6 & 12.7$\pm$2.2 & 46.9$\pm$1.4 & \textbf{17.2} \\
 & LoRA (FedAvg) & 0.3 M & 15.9  & 0.0 & -2.0  & 3.3 & -0.1 & 15.8 &  5.48 \\
 & VeFA (FedAvg) & 0.018 M & 4.7  & 0.0  & -4.0 & 5.7 &  1.1 & 43.7 & \textbf{8.53} \\
 & LoRA ($\bar{R}$) & 0.3 M & 7.16 & 18.2 & 4.9 & -4.1 & 4.3 & 8.6 & 6.5 \\
 & VeFA ($\bar{R}$) & 0.018 M & 13.38 & 9.3 & -1.6 & 4.4 & 11.2 & 31.2 & \textbf{11.3} \\
\cmidrule(lr){1-10}
\multirow{8}{*}{\rotatebox{90}{\textbf{Large}}} & LoRA ($\tau=1$) & 0.3 M & -1.2$\pm$1.4& 0& 0& -4.6$\pm$1.3& -0.6$\pm$0.6& 15.4$\pm$9.1 & 1.5 \\
&  VeFA ($\tau=1$) & 0.018 M & -0.6$\pm$3.1& 19.6$\pm$12.4 & 6.6$\pm$1.7 & 15.8$\pm$2.5 & 18.2$\pm$2.0 & 60.4$\pm$4.0 & \textbf{20.0} \\
& LoRA ($\tau=10$)& 0.3 M & -1.9$\pm$0.6 & 0 & 0 & 0.2$\pm$0.1 & -0.3$\pm$1.6& 9.7$\pm$1.1 & 1.3 \\
& VeFA ($\tau=10$)& 0.018 M & -1.0$\pm$0.6 & 21.1$\pm$11.2 & 5.4$\pm$0.9 & 15.4$\pm$1.5 & 19.0$\pm$3.7 & 57.9$\pm$4.5 & \textbf{19.6}\\
 & LoRA (FedAvg) & 0.3 M & -2.0  & 0.0 & 0.0 & 0.1 & -1.1 & 34.0 & 5.17 \\
 & VeFA (FedAvg) & 0.018 M & 11.0 & 6.8 & 14.1 & 8.7 & 11.2 & 59.0 & \textbf{18.47} \\
  & LoRA ($\bar{R}$) & 0.3 M & 2.26 & 12.1 & 4.8 & 3.8 & 4.1 & 18.2 & 7.5 \\
 & VeFA ($\bar{R}$) & 0.018 M & 8.90 & 23.4 & 2.6 & 9.1 & 12.8 & 55.3 & \textbf{18.7} \\
\bottomrule
\end{tabular}
\end{footnotesize}
\end{table}

\subsection{Maximum VeFA Scaling Coefficient}
\label{appendix: Maximum VeFA Scaling Coefficient}
Tables \ref{tab:max-k-img}, \ref{tab:max-k-nlu} and \ref{tab:max-k-nlg} report the maximum VeFA scaling coefficient $\|\boldsymbol{k}\|_\infty$ across all layers for each task and backbone. For image classification (Table \ref{tab:max-k-img}), all datasets yield $\|\boldsymbol{k}\|_\infty \leq 1.9$, indicating only mild rescaling of feature dimensions. On GLUE NLU tasks (Table  \ref{tab:max-k-nlu}), both RoBERTa-Base and RoBERTa-Large exhibit similarly small scaling factors, mostly in the range 1.4–3.9. For NLG with GPT-2 (Table \ref{tab:max-k-nlg}), the maximal values remain below 6.2. Overall, these results confirm that VeFA operates with relatively small feature-wise scalings in practice, which is consistent with our theoretical analysis that the forgetting under VeFA is bounded and controlled by $\|\boldsymbol{k}\|_\infty$.

\begin{table}[H]
\centering
\caption{Maximum VeFA scaling coefficient $\|\boldsymbol{k}\|_\infty$ on image classification tasks.}
\label{tab:max-k-img}
\footnotesize
\setlength{\tabcolsep}{6pt}
\renewcommand{\arraystretch}{1.2}
\begin{tabular}{lccccccc}
\toprule
 & UCF101 & Oxford Pets & Oxford Flowers & EuroSAT & DTD & Caltech101 & Food101 \\
\midrule
$\|\boldsymbol{k}\|_\infty$
 & 1.766 & 1.558 & 1.568 & 1.120 & 1.855 & 1.665 & 1.460 \\
\bottomrule
\end{tabular}
\end{table}

\begin{table}[H]
\centering
\caption{Maximum VeFA scaling coefficient $\|\boldsymbol{k}\|_\infty$ on GLUE NLU tasks.}
\label{tab:max-k-nlu}
\footnotesize
\setlength{\tabcolsep}{6pt}
\renewcommand{\arraystretch}{1.2}
\begin{tabular}{lcccccc}
\toprule
 & CoLA & MRPC & QNLI & RTE & SST-2 & STS-B \\
\midrule
RoBERTa-Base  & 3.889 & 1.879 & 3.256 & 2.693 & 1.522 & 1.527 \\
RoBERTa-Large & 3.028 & 1.671 & 3.061 & 1.886 & 1.544 & 1.404 \\
\bottomrule
\end{tabular}
\end{table}

\begin{table}[H]
\centering
\caption{Maximum VeFA scaling coefficient $\|\boldsymbol{k}\|_\infty$ on NLG tasks.}
\label{tab:max-k-nlg}
\footnotesize
\setlength{\tabcolsep}{6pt}
\renewcommand{\arraystretch}{1.2}
\begin{tabular}{lcc}
\toprule
 & GPT-2-Medium & GPT-2-Large \\
\midrule
$\|\boldsymbol{k}\|_\infty$ & 2.0 & 2.0 \\
\bottomrule
\end{tabular}
\end{table}

\subsection{Analysis of Intruder Dimensions in LoRA}
\label{additional_analysis_lora_intruder}
\begin{figure}[htbp]
    \centering
    \includegraphics[width=0.6\textwidth]{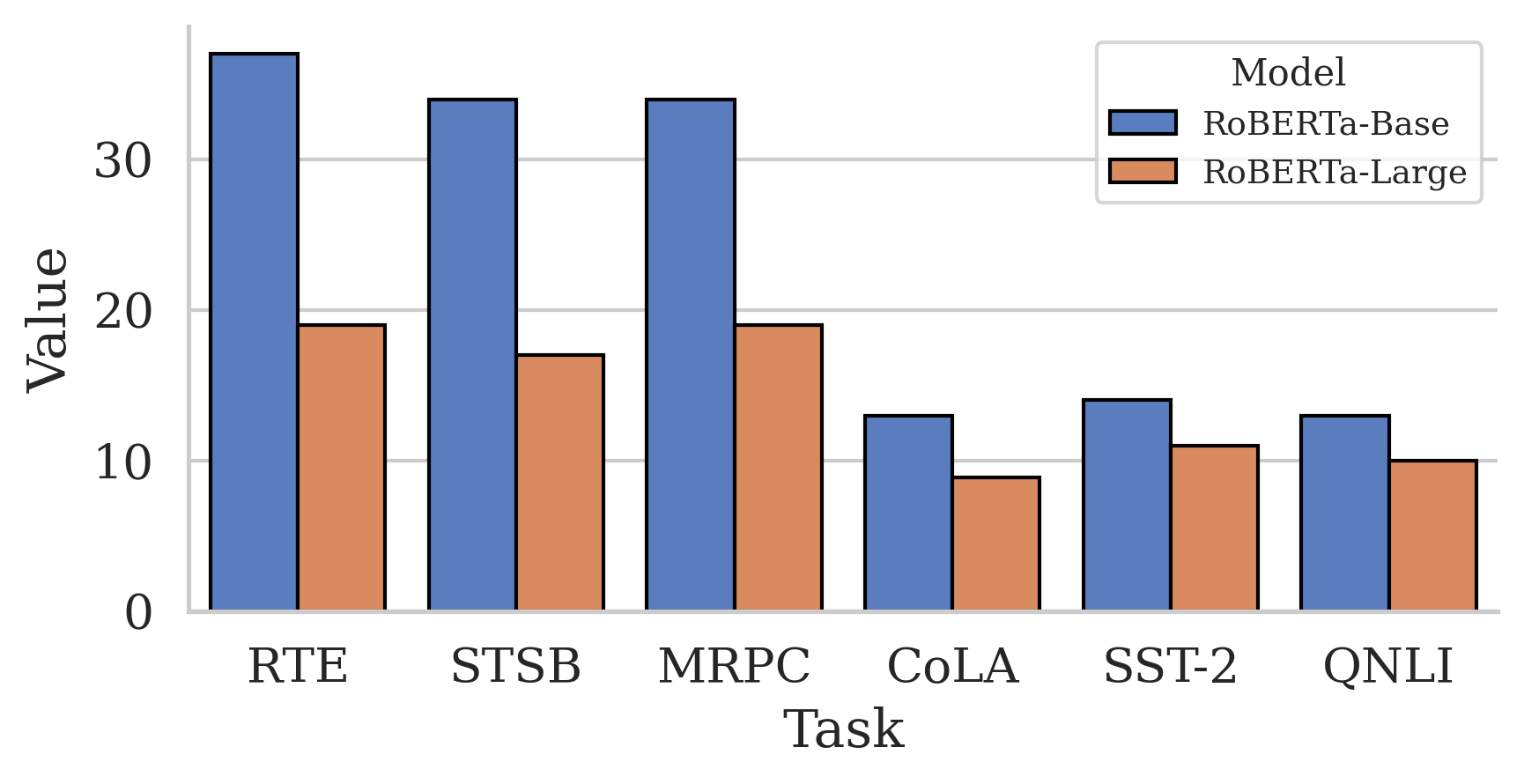}
    \caption{Number of intruder dimensions for different GLUE tasks.}
    \label{fig: number_intruder_dimension}
\end{figure}

\begin{figure}[t]
    \centering
    \begin{subfigure}[t]{0.45\linewidth}
        \centering
        \includegraphics[width=\linewidth]{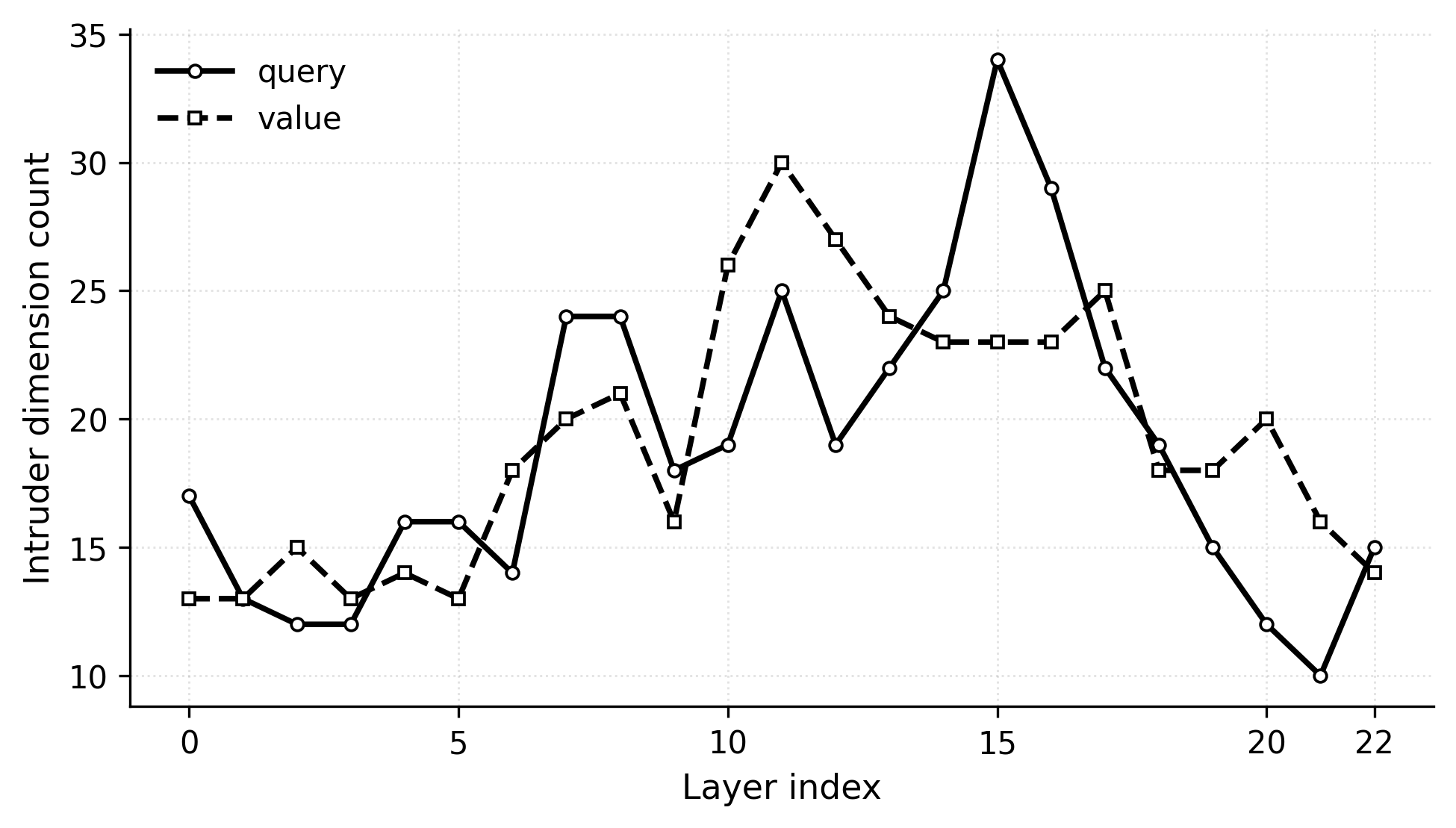}
        \caption{RTE}
    \end{subfigure}
    \hfill
    \begin{subfigure}[t]{0.45\linewidth}
        \centering
        \includegraphics[width=\linewidth]{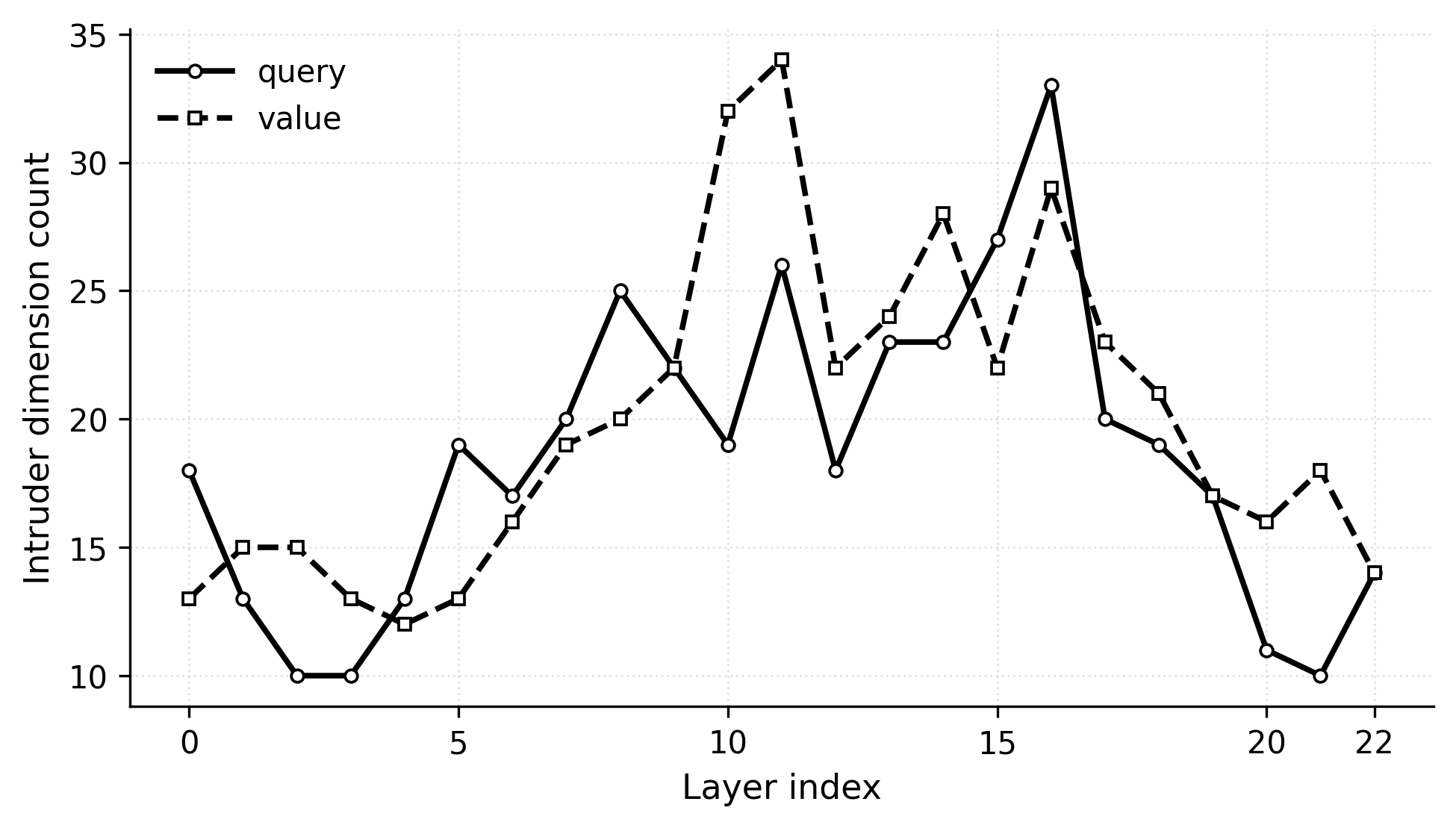}
        \caption{MRPC}
    \end{subfigure}
    \caption{Number of intruder dimensions in LoRA query and value matrices across layers of RoBERTa-L.}
    \label{fig: number_intruder_LoRA}
\end{figure}

We firstly report the number of intruder dimensions for different GLUE tasks in Fig.\ref{fig: number_intruder_dimension}. It can be seen that LoRA consistently introduces intruder dimensions across tasks, which may exacerbate forgetting and cause conflicts when merging adapters, as shown in Tab.\ref{Tab: GLUE_robustness}. This also explains why, in our cross-task evaluations, LoRA underperforms VeFA, which introduces no additional intruder dimensions. In particular, RTE, STS-B, and MRPC exhibit more intruder dimensions than the other three tasks, likely because they are all fine-tuned from the same MNLI checkpoint. A layer-wise analysis can be seen in Fig. \ref{fig: number_intruder_LoRA}. Intruder directions are present across most layers, with a pronounced concentration in mid-to-late layers for both RTE and MRPC. Moreover, query and value exhibit comparable intruder counts, but their relative dominance varies by layer, indicating that intruder behavior is module- and layer-specific rather than uniformly attributable to a single projection.

\begin{figure}[t]
    \centering
    \begin{subfigure}[t]{0.45\linewidth}
        \centering
        \includegraphics[width=\linewidth]{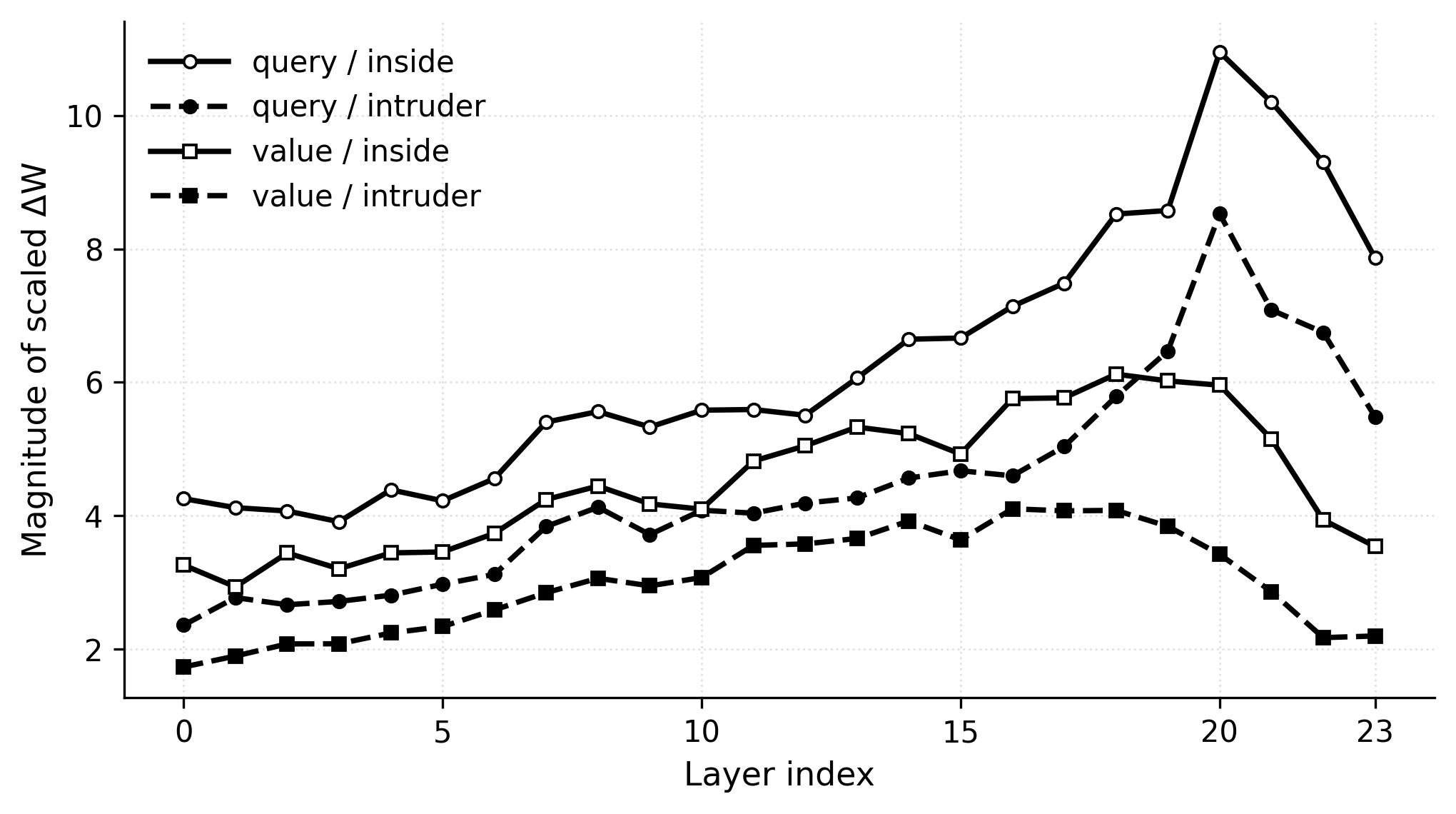}
        \caption{RTE}
    \end{subfigure}
    \hfill
    \begin{subfigure}[t]{0.45\linewidth}
        \centering
        \includegraphics[width=\linewidth]{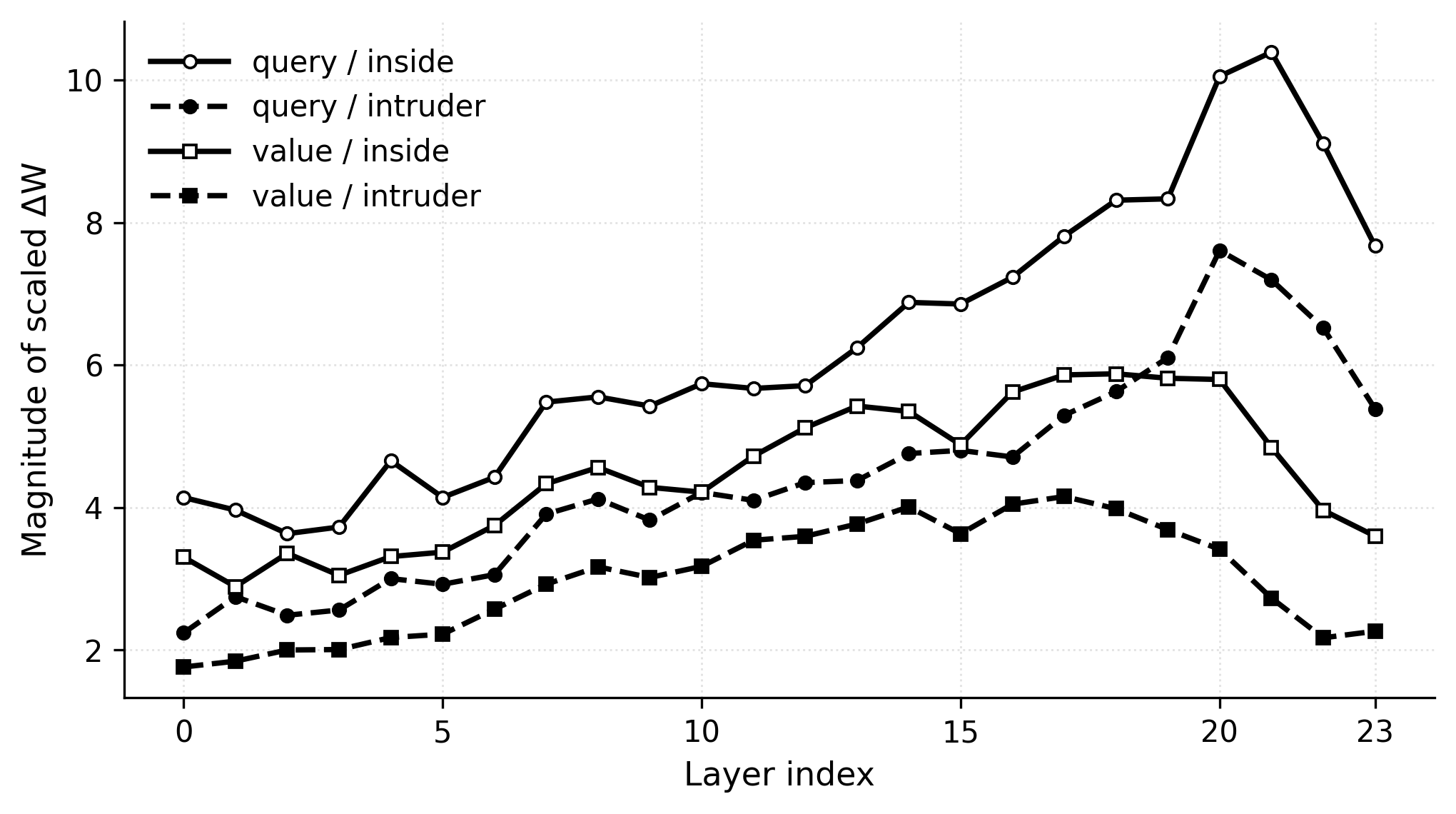}
        \caption{MRPC}
    \end{subfigure}
    \caption{Magnitude of the adapted $\boldsymbol{\Delta}$ vectors in LoRA query and value matrices across layers of RoBERTa-L.}
    \label{fig: magnitude_adaptation_LoRA}
\end{figure}

Building on the layer-wise statistics of intruder-dimension counts (Fig. \ref{fig: number_intruder_LoRA}), we further conduct a finer-grained analysis by decomposing the LoRA update $\boldsymbol{\Delta}$ into two components: (i) the part that lies inside the column space of the pretrained weight $\boldsymbol{W}_0$, and (ii) the remaining component residing in the intruder subspace. As shown in Fig. \ref{fig: magnitude_adaptation_LoRA}, for both RTE and MRPC, the “inside” component consistently dominates the overall update magnitude, while the intruder component is non-negligible and becomes more pronounced in later layers, especially for the query matrices. Notably, the trends in intruder counts and intruder magnitudes are not strictly aligned: a layer may exhibit a moderate number of intruder dimensions yet carry substantial intruder mass (or vice versa), suggesting that “how many” intruder directions appear and “how much” update energy they carry capture complementary aspects of the adaptation behavior.

\end{document}